\documentclass{article}

\usepackage[final]{neurips_2025}

\usepackage{amsmath,amsfonts,bm}

\def\eqref#1{equation~\ref{#1}}

\def\1{\bm{1}}

\DeclareMathAlphabet{\mathsfit}{\encodingdefault}{\sfdefault}{m}{sl}
\SetMathAlphabet{\mathsfit}{bold}{\encodingdefault}{\sfdefault}{bx}{n}

\def\sR{{\mathbb{R}}}

\usepackage[utf8]{inputenc} 
\usepackage[T1]{fontenc}    
\usepackage{hyperref}       
\usepackage{url}            
\usepackage{booktabs}       
\usepackage{amsfonts}       
\usepackage{nicefrac}       
\usepackage{microtype}      
\usepackage{xcolor}         
\usepackage{graphicx}
\usepackage{multirow}
\usepackage{stmaryrd}
\usepackage{amssymb}
\usepackage{dsfont}

\title{Spectral Convolutional Conditional Neural Processes}

\author{%
  Peiman Mohseni \\
  Texas A\&M University \\
  \texttt{peiman.mohseni@tamu.edu} \\
  \And
  Nick Duffield \\
  Texas A\&M University \\
  \texttt{duffieldng@tamu.edu} \\
}

\begin{document}

\maketitle

\begin{abstract}
Neural Processes (NPs) are meta-learning models that learn to map sets of observations to approximations of the corresponding posterior predictive distributions. By accommodating variable-sized, unstructured collections of observations and enabling probabilistic predictions at arbitrary query points, NPs provide a flexible framework for modeling functions over continuous domains. Since their introduction, numerous variants have emerged; however, early formulations shared a fundamental limitation: they compressed the observed data into finite-dimensional global representations via aggregation operations such as mean pooling. This strategy induces an intrinsic mismatch with the infinite-dimensional nature of the stochastic processes that NPs intend to model. Convolutional conditional neural processes (ConvCNPs) address this limitation by constructing infinite-dimensional functional embeddings processed through convolutional neural networks (CNNs) to enforce translation equivariance. Yet CNNs with local spatial kernels struggle to capture long-range dependencies without resorting to large kernels, which impose significant computational costs. To overcome this limitation, we propose spectral ConvCNPs (SConvCNPs), which perform global convolution in the frequency domain. Inspired by Fourier neural operators (FNOs) for learning solution operators of partial differential equations (PDEs), our approach directly parameterizes convolution kernels in the frequency domain, leveraging the relatively compact yet global Fourier representation of many natural signals. We validate the effectiveness of SConvCNPs on both synthetic and real-world datasets, demonstrating how ideas from operator learning can advance the capabilities of NPs.
\end{abstract}

\section{Introduction}\label{sec: intro}

Stochastic processes offer a mathematical framework for modeling systems that evolve with inherent randomness over continuous domains such as time and space. They underpin a wide range of scientific applications---from spatio-temporal climate dynamics to biological and physical systems---thereby motivating the development of machine learning methods that can learn from data generated by such phenomena \citep{ mathieu2021contrastive, vaughan2021convolutional, allen2025end, ashman2025gridded, dupont2021generative}. Among classical approaches, Gaussian processes (GPs; \citealp{rasmussen2006gaussian}) provide a Bayesian framework with closed-form inference and uncertainty quantification. However, their cubic computational cost from matrix inversion and the difficulty of specifying suitable kernels---especially in high-dimensional settings---limit their scalability.

Motivated by the success of deep neural networks in large-scale function approximation, neural network-based alternatives have emerged. Neural processes (NPs; \citealp{garnelo2018conditional, garnelo2018neural}) exemplify this paradigm, combining ideas from GPs and deep learning within a meta-learning framework. By exposing the model to multiple realizations of an underlying stochastic process, each treated as a distinct task, NPs learn shared structures across tasks to parameterize a neural mapping that \emph{directly} gives an approximation of the corresponding posterior predictive distribution \citep{bruinsma2024convolutional}. Once trained, the model enables efficient probabilistic predictions on new tasks without further training.

Since the introduction of conditional neural processes (CNPs, \citet{garnelo2018conditional}) as the first class within the NPs family \citep{jha2022neural}, numerous extensions have been proposed to enhance their effectiveness. One prominent line of work focuses on incorporating explicit inductive biases into CNPs in order to better capture the symmetries that commonly arise in scientific applications \citep{gordon2019convolutional, kawano2021group, holderrieth2021equivariant, huang2023practical, ashman2024translation, ashman2024approximately}. 
Another major direction seeks to move beyond the mean-field factorized Gaussian predictive distributions to which CNPs are limited. A widely adopted approach augments CNPs with stochastic latent variables, giving rise to the family of latent neural processes (LNPs; \citet{garnelo2018neural, louizos2019functional, wang2020doubly, foong2020meta, lee2020bootstrapping, volpp2021bayesian, wang2022bridge, wang2022learning, kim2022neural, jung2023bayesian, lee2023martingale, xu2023deep}).
Complementary efforts explore autoregressive prediction schemes \citep{bruinsma2023autoregressive, nguyen2022transformer}, Gaussian predictive distributions with non-diagonal covariances \citep{bruinsma2021gaussian, markou2022practical}, and quantile-based parameterizations of the predictive distribution \citep{mohseni2023adaptive}.

This work focuses on CNPs, particularly convolutional CNPs (ConvCNPs; \citet{gordon2019convolutional}), which were the first to endow NPs with translation equivariance. ConvCNPs introduce the convolutional deep set construction which characterizes a broad class of translation-equivariant mappings over finite, potentially unstructured sets of observations as a composition of a functional embedding with a translation-equivariant operator, typically realized through convolutional neural networks (CNNs; \citet{fukushima1980neocognitron, lecun1989backpropagation, lecun1998gradient}).

Despite their effectiveness, ConvCNPs can struggle to aggregate information from observations spread across large spatial domains---a challenge that becomes particularly pronounced in sparse data regimes. This limitation stems from their reliance on local convolutional kernels with small receptive fields, which hampers their ability to model long-range dependencies. A natural remedy is to enlarge the kernel size to extend the receptive field; however, this approach rapidly increases the number of model parameters and computational cost \citep{romero2021ckconv}. Alternatively, transformer-based architectures can capture long-range interactions but incur quadratic, rather than linear, computational complexity in the number of observations \citep{vaswani2017attention, nguyen2022transformer}.

In this work, we pursue an alternative paradigm that represents functions in the frequency domain, inspired by the well-established observation that many natural processes exhibit energy concentration in low-frequency bands \citep{field1987relations, ruderman1993statistics, wainwright1999scale}. This property allows for efficient approximation using only a subset of dominant spectral coefficients, enabling tractable computations while preserving the signal’s global structure. By parameterizing convolution kernels directly in the Fourier domain over a finite set of frequencies and leveraging the convolution theorem, we can attain large effective receptive fields without incurring prohibitive computational costs.

While spectral methods have been extensively studied in neural operator learning for partial differential equations (PDEs, \citet{chen1995universal, kovachki2021universal, kovachki2023neural}), their application within NPs framework remains relatively unexplored. To bridge this gap, we propose spectral convolutional conditional neural processes (SConvCNPs)—a class of models that adopt Fourier neural operators (FNOs; \citet{li2020fourier}) to realize global spectral convolution while maintaining computational efficiency. Across a suite of synthetic and real-world benchmarks, SConvCNPs perform competitively with state-of-the-art baselines, illustrating how ideas from neural operators can enhance the flexibility and performance of NPs.

\section{Preliminaries}\label{sec: prelim}
\subsection{Fourier Neural Operators}\label{subsec: FNOs}

Neural operators~\citep{chen1995universal, li2020fourier, kovachki2023neural} are neural network architectures designed to learn mappings between \emph{function spaces} rather than finite-dimensional vectors. Like conventional feed-forward networks, they consist of stacked layers that alternate between operator-based transformations and pointwise nonlinearities. A common transformation is an integral operator with kernel $\kappa : \mathcal{X} \times \mathcal{X} \to \mathcal{Y}$, acting on an input function $v : \mathcal{X} \to \mathcal{Y}$ as
\begin{equation*}
    \mathcal{K}[v](x) = \int \kappa(x, s)\, v(s)\, \mathrm{d}s,
\end{equation*}
where $\mathcal{X}=\sR^{d_x}$ and $\mathcal{Y}=\sR^{d_y}$ are Euclidean spaces with $d_x, d_y\in \mathbb{N}$. While this work focuses on operators linear in $v$, nonlinear formulations have also been explored, including continuous formulations of softmax attention \citep{ashman2024translation, calvello2024continuum}. 
When the kernel is \emph{stationary}---that is, $\kappa(x, s) = \kappa(x - s)$---the operator reduces to a convolution, $\mathcal{K}[v] = \kappa \ast v$, connecting neural operators to CNNs where $\kappa$ is parameterized by learnable weights.

Convolutional kernels are typically spatially local with limited receptive fields \citep{luo2016understanding, Peng_2017_CVPR, wang2018non}. Modeling long-range dependencies thus requires large kernels, substantially increasing parameter count. The Fourier Neural Operator (FNO;~\citet{li2020fourier}) addresses this limitation by exploiting the convolution theorem~\citep{bracewell1966fourier, oppenheim1999discrete}, which expresses convolution as
\vspace{-0.2cm}
\begin{equation}\label{eq: convolution-theorem}
    \mathcal{K}[v](x) = \mathcal{F}^{-1} \Big[ \widehat{\kappa}(\xi) \cdot \widehat{v}(\xi) \Big](x),
\end{equation}
where $\widehat{f} := \mathcal{F}[f]$ denotes the Fourier transform of $f$, and $\mathcal{F}^{-1}$ denotes the inverse Fourier transform. Rather than parameterizing $\kappa$ in the spatial domain, the FNO learns its Fourier representation $\widehat{\kappa}$ directly. When $v$ is approximately band-limited---i.e., $\widehat{v}(\xi)$ has negligible energy for $\|\xi\| > \xi_0$---high-frequency components can be truncated with minimal information loss. This property, observed in many natural signals~\citep{field1987relations, ruderman1993statistics, wainwright1999scale}, allows setting $\widehat{\kappa}(\xi) = 0$ outside the retained band without significant degradation.

In practice, functions are accessible only through discrete samples, requiring the discrete Fourier transform (DFT) for domain transitions. Given samples of $v$ on a \emph{uniform} grid $\mathcal{G} \subset \mathcal{X}$, the FNO applies the DFT via the fast Fourier transform (FFT;~\citet{cooley1965algorithm, frigo2005design}). The resulting spectrum is truncated to a finite set of frequency modes $\Xi \subset \mathbb{R}^{d_x}$ assumed to capture most of the signal's energy. For each retained frequency $\xi \in \Xi$, the Fourier kernel is parameterized by learnable complex weights $W_\xi \in \mathbb{C}^{d_y}$, so that $\widehat{\kappa}(\xi) = W_\xi$.\footnote{More generally, a matrix-valued parameterization $W_\xi \in \mathbb{C}^{c_{\text{out}} \times d_y}$ is used, where $c_{\text{out}}$ is the number of output channels, enabling joint mixing of input channels and projection to different dimensionality.} This formulation implicitly imposes spatial periodicity, as the kernel is represented using discrete harmonics (equivalently, a Dirac comb in frequency space). After pointwise multiplication in frequency space, an inverse FFT returns the operator output to the spatial domain.

\subsection{Neural Processes}\label{subsec: NPs}
Let $\mathcal{P}(\mathcal{Y}^{\mathcal{X}})$ denote the space of probability measures over measurable functions $f:\mathcal{X}\to\mathcal{Y}$, which we interpret as stochastic processes indexed by $\mathcal{X}$. We assume an unknown data-generating process $\mu \in \mathcal{P}(\mathcal{Y}^{\mathcal{X}})$, from which latent functions are drawn. A \emph{task} consists of a \emph{finite} number of noisy input--output observations generated from a realization $f \sim \mu$, partitioned into a context set and a query set:
\begin{equation}\label{eq: task}
\mathcal{D} = (\mathcal{D}_c, \mathcal{D}_q),
\qquad
\mathcal{D}_c = \{(x_{c,k}, y_{c,k})\}_{k=1}^{n_c},
\qquad
\mathcal{D}_q = \{(x_{q,l}, y_{q,l})\}_{l=1}^{n_q}.
\end{equation}
Observations are generated according to
\[
y_{c,k} = f(x_{c,k}) + \epsilon_{c,k},
\qquad
y_{q,l} = f(x_{q,l}) + \epsilon_{q,l},
\]
where $\epsilon_{c,k}, \epsilon_{q,l} \stackrel{\text{i.i.d.}}{\sim} \mathcal{N}(0,\sigma_0^2)$ and $\sigma_0>0$ denotes the observation noise standard deviation.

Neural processes (NPs; \citep{garnelo2018conditional, garnelo2018neural}) constitute a class of models that use neural networks to learn a mapping
$\eta:\mathcal{S}(\mathcal{X}\times\mathcal{Y}) \to \mathcal{P}(\mathcal{Y}^{\mathcal{X}})$, where $\mathcal{S}(\mathcal{X}\times\mathcal{Y})$ denotes the collection of all \emph{finite} subsets of $\mathcal{X}\times\mathcal{Y}$. Given a context set $\mathcal{D}_c$, $\eta$ outputs a stochastic process intended to approximate the Bayesian posterior over functions induced by $\mu$ and conditioned on $\mathcal{D}_c$. Typically, this process is specified \emph{implicitly} through its finite-dimensional marginals \citep{garnelo2018neural, bruinsma2021gaussian, bruinsma2024convolutional, mathieu2023geometric}.
Let $\mathbf{x}_q = (x_{q,l})_{l=1}^{n_q}$ and $\mathbf{y}_q = (y_{q,l})_{l=1}^{n_q}$. We denote by $\eta[\mathcal{D}_c; \mathbf{x}_q](\cdot)$ the finite-dimensional distribution of the process $\eta[\mathcal{D}_c]$ evaluated at $\mathbf{x}_q$, and by $\mu_{\mathbf{x}_q}(\cdot \mid \mathcal{D}_c)$ the corresponding finite-dimensional marginal of the true posterior process.
Informally, the NP approximation aims to satisfy
\[
\eta[\mathcal{D}_c; \mathbf{x}_q](\cdot)
\;\approx\;
\mu_{\mathbf{x}_q}(\cdot \mid \mathcal{D}_c),
\]
for all $\mathbf{x}_q$ and $\mathcal{D}_c$ \citep{bruinsma2021gaussian, bruinsma2024convolutional}. With slight abuse of notation, we write $\eta[\mathcal{D}_c; \mathbf{x}_q](\mathbf{y}_q)$ for the density of this finite-dimensional distribution evaluated at $\mathbf{y}_q$.

In this work, we focus on \emph{conditional neural processes} (CNPs; \citet{garnelo2018conditional}), which restrict these finite-dimensional distributions to mean-field Gaussians, i.e. $\eta[\mathcal{D}_c; \mathbf{x}_q] = \prod_{l=1}^{n_q}\eta[\mathcal{D}_c; x_{q,l}]$, where each marginal $\eta[\mathcal{D}_c; x_{q,l}]$ is Gaussian. Each predictive marginal is typically parameterized via a two-stage encoder--decoder architecture \citep{bruinsma2024convolutional, ashman2024translation, ashman2025gridded}.
The encoder $\varphi_e:\mathcal{S}(\mathcal{X}\times\mathcal{Y}) \to \mathcal{H}$ maps the context set $\mathcal{D}_c$ to a latent representation, while the decoder $\varphi_d: \mathcal{H} \to \Theta^{\mathcal{X}}$ maps this representation to a function that assigns, to each query $x_{q,l}$, parameters $\theta(x_{q,l}) \in \Theta$ of the predictive distribution $\eta[\mathcal{D}_c; x_{q,l}]$.

The vanilla CNP encoder summarizes $\mathcal{D}_c$ using a \emph{permutation-invariant} architecture \citep{qi2017pointnet, zaheer2017deep}. Each context pair $(x_{c,k}, y_{c,k})\in\mathcal{D}_c$ is independently mapped by a shared network to a finite-dimensional embedding $\varepsilon_{c,k}$, which are then aggregated---typically via mean pooling---into a single representation $\varepsilon_c$. Notably, this encoding is independent of the query locations. The decoder then combines $\varepsilon_c$ with each $x_{q,l}$ to parameterize the Gaussian predictive distribution.

Although sum-pooling aggregation provides universal approximation guarantees \citep{zaheer2017deep, bloem2020probabilistic}, NPs employing such mechanisms often exhibit underfitting in practice \citep{kim2019attentive}. Prior works have partly attributed this phenomena to two primary factors \citep{xu2020metafun}: (1) the limitation of summaries with \emph{prespecified finite dimensionality} in representing context sets of arbitrary size \citep{wagstaff2019limitations}, and (2) the shortcomings of simple sum or mean pooling operations to effectively capture rich dependencies between different points \citep{xu2020metafun, nguyen2022transformer}.

Since NPs address inherently functional learning problems, it is natural to consider embeddings that themselves take the form of functions. \citet{gordon2019convolutional} introduce a framework for translation-equivariant prediction maps over sets,
satisfying
\[
\eta\Big[\{(x_c+\tau, y_c)\mid(x_c,y_c)\in\mathcal{D}_c\}; \mathbf{x}_q\Big]
=
\eta\big[\mathcal{D}_c; \mathbf{x}_q-\tau\big],
\]
for all translations $\tau$, context sets $\mathcal{D}_c$, and collections of query locations $\mathbf{x}_q$, where subtraction by $\tau$ acts pointwise, i.e., $\mathbf{x}_q-\tau = (x_{q,l}-\tau)_{l=1}^{n_q}$. They show that a broad class of such maps can be written as $\eta[\mathcal{D}_c] = \varphi_d \circ \varphi_e[\mathcal{D}_c]$, where the functional embedding is defined by
\begin{equation}\label{eq: functional-embedding}
    \varphi_e[\mathcal{D}_c](x)
    = \mkern-15mu
    \sum_{(x_c,y_c)\in\mathcal{D}_c} \mkern-15mu \phi(y_c)\,\psi_e(x-x_c).
\end{equation}
Moreover, $\varphi_d:\mathcal{H}\to C_b(\mathcal{X},\mathcal{Y})$ is a translation-equivariant decoder acting on a function space $\mathcal{H}$, $C_b(\mathcal{X},\mathcal{Y})$ denotes the space of bounded continuous functions from$\mathcal{X}$ to $\mathcal{Y}$, $\phi(y)=(1,y)$%
\footnote{More generally, $\phi(y)=(1,y,\dots,y^M)$, where $M$ accounts for repeated inputs.See \citet{gordon2019convolutional} and \citet{bruinsma2024convolutional} for details.}, and $\psi_e:\mathcal{X}\to\mathbb{R}$ is a continuous strictly positive-definite kernel, typically Gaussian.

In convolutional conditional neural processes (ConvCNPs), the functional embedding $\varphi_e[\mathcal{D}_c]$ is evaluated on a uniform grid $\mathcal{G}\subset\mathcal{X}$ covering the joint support of $\mathbf{x}_c$ and $\mathbf{x}_q$, yielding the discretized representation $(\varphi_e[\mathcal{D}_c](x_g))_{x_g\in\mathcal{G}}$. This representation is processed by $\varphi_d$, and predictions at query locations are obtained via kernel interpolation:
\begin{equation}\label{eq: ConvCNP-decoder-interpolate}
\theta_{q,l}
=
\sum_{x_g\in\mathcal{G}}
(\varphi_d \circ \varphi_e )[\mathcal{D}_c](x_g)\,\psi_d(x_{q,l}-x_g),
\qquad l=1,\dots,n_q,
\end{equation}
where $\psi_d$ is another strictly positive-definite kernel. This interpolation step may be viewed as part of the decoder, preserving the overall encoder--decoder abstraction.

\section{Spectral Convolutional Conditional Neural Processes}\label{sec: method}
The decoder $\varphi_d$ in ConvCNPs is typically parameterized using standard CNNs such as U-Net~\citep{ronneberger2015u} or ResNet~\citep{he2016deep}. These architectures employ discrete convolutional kernels---finite sets of learnable parameters that define localized filters operating over neighboring grid points. The kernel size, fixed \emph{a priori}, determines the receptive field of each convolution \citep{ding2022scaling} and is generally much smaller than the overall extent of the input signals in the physical domain~\citep{romero2021ckconv, knigge2023modelling}.
This locality constraint fundamentally limits a model’s ability to capture long-range dependencies and to integrate information from observations distributed across large spatial or temporal domains \citep{Peng_2017_CVPR, wang2018non, ramachandran2019stand, wang2020axial}. 
The issue becomes particularly pronounced when handling sparse or irregularly sampled data, where effective global reasoning cannot emerge solely from local convolution operations.
Although enlarging the convolutional kernel increases the receptive field, it results in a rapid escalation of both parameter count and computational cost.
Transformer-based architectures mitigate this issue by enabling explicit global interactions \citep{vaswani2017attention}; however, they incur quadratic, rather than linear, complexity in the number of inputs, rendering them impractical for large context sets unless one resorts to approximation schemes \citep{nguyen2022transformer, feng2022latent, ashman2024translation, ashman2025gridded}.


To overcome this limitation without relying on prohibitively large filters or expensive transformers, we exploit the Fourier representation of signals. This is motivated by the well-established observation that many natural signals are approximately band-limited (see Section~\ref{subsec: FNOs}), implying that their Fourier representation offers a more compact encoding \emph{relative} to its physical-domain counterpart. Concretely, we instantiate $\varphi_d$ via spectral convolution modules based on \eqref{eq: convolution-theorem}.  This substitution effectively enlarges the receptive field, enabling the model to capture global structures from sparse or irregularly sampled data without incurring a parameter count explosion. We refer to the resulting models as spectral convolutional conditional neural processes (SConvCNPs).

\paragraph{Computational Complexity.}
The computational cost of SConvCNPs comprises three parts:
(i) $\mathcal{O}(|\mathcal{D}_c||\mathcal{G}|)$ to compute the discretized functional embedding on the grid $\mathcal{G}$ (\eqref{eq: functional-embedding}); 
(ii) $\mathcal{O}(|\mathcal{G}|\log |\mathcal{G}|)$ for the FFT-based spectral convolutions (\eqref{eq: convolution-theorem}); and 
(iii) $\mathcal{O}(|\mathcal{D}_q||\mathcal{G}|)$ to interpolate grid embeddings at the query locations (\eqref{eq: ConvCNP-decoder-interpolate}). 
Overall, the total complexity is $\mathcal{O}\bigl(|\mathcal{G}| (|\mathcal{D}_c| + \log|\mathcal{G}| + |\mathcal{D}_q|)\bigr)$, matching the $\mathcal{O}\bigl(|\mathcal{G}| (|\mathcal{D}_c| + 1 + |\mathcal{D}_q|)\bigr)$ complexity of ConvCNPs up to logarithmic factors. Both architectures therefore scale \emph{linearly} with task size. A key limitation, however, is that $|\mathcal{G}|$ grows exponentially with the input dimension $d_x$, restricting these methods to low-dimensional domains. In contrast, transformer-based NPs (TNPs; \citet{kim2019attentive, nguyen2022transformer, feng2022latent, ashman2024translation}) avoid gridding entirely and thus scale more gracefully with input dimensionality. Instead, their computational cost scales quadratically with the task size,
typically as
$\mathcal{O}(|\mathcal{D}_c|^2 + |\mathcal{D}_q|^2)$,
with minor variations depending on the specific implementation
(see Section~\ref{appendix: synthetic-regression-model-architectures}).

\paragraph{Positional Encodings}
The convolution operator preserves translation equivariance under the Fourier transform (see \eqref{eq: convolution-theorem}). However, practical FNOs implementations often include explicit positional information to improve predictive accuracy~\citep{li2020fourier, tran2021factorized, gupta2021multiwavelet, rahman2022uno, helwig2023group, tripura2023wavelet, liu2023domain, li2024multi}. Accordingly, we augment the functional embedding $\varphi_e[\mathcal{D}_c](x)$ with positional information:
\[
\widetilde{\varphi}_e[\mathcal{D}_c](x)
= \big(\varphi_e[\mathcal{D}_c](x),\, x\big).
\]
While this augmentation explicitly breaks translation equivariance, we empirically observe performance improvements (see Section~\ref{appendix: ablation}).
An interesting direction for future work is to investigate \emph{relative} positional encodings
\citep{shaw2018self, su2024roformer}, which can provide spatial context while preserving translation equivariance.

\paragraph{Discretization Sensitivity of DFT.}
Unlike the continuous Fourier transform, the DFT—and therefore the FFT—depends inherently on the grid $\mathcal{G}$ on which $\varphi_e[\mathcal{D}_c]$ (or $\widetilde{\varphi}_e[\mathcal{D}_c]$) is represented. This dependence arises from both the grid resolution and its physical extent: changing either alters the resulting Fourier coefficients and can lead to inconsistent behavior in the outputs (see Section \ref{appensix: DFT-discretization}). Sensitivity to resolution is not unique to DFT; CNNs exhibit analogous issues \citep{raonic2023convolutional, bartolucci2023representation}. For example, ConvCNPs mitigate this effect by fixing the grid resolution. While spatial CNNs become stable once the resolution is fixed, DFT-based methods remain sensitive unless \emph{both} the resolution and the physical range are controlled. Accordingly, we fix both, choosing a domain sufficiently large to cover all context and query inputs across tasks. When this is impractical, the domain can instead be divided into (possibly overlapping) fixed-size patches, with spectral convolutions applied independently to each patch and the outputs aggregated—mirroring the mechanism of standard convolution layers. Related ideas in efficient transformer architectures suggest that this is a promising direction for future work~\citep{beltagy2020longformer, zaheer2020big, liu2021swin, ding2023longnet}.

\section{Experiments}\label{sec: experiments}
\begin{figure*}
    \centering
    \hspace*{-0.1cm} 
    \scalebox{1.0}{
        \begingroup
        \setlength{\tabcolsep}{-1pt}
        \begin{tabular}{l@{\hskip -0.2pt}ccc}
            & &
            \makebox[0pt][c]{%
                    \includegraphics[width=0.5\textwidth]{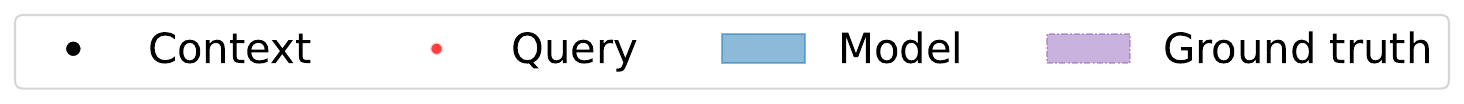}%
            }
            &\\
            \raisebox{3.85\normalbaselineskip}[0pt][0pt]{\rotatebox[origin=c]{90}{\scriptsize GP-Matérn 5/2}} &
            \includegraphics[width=0.33\textwidth]{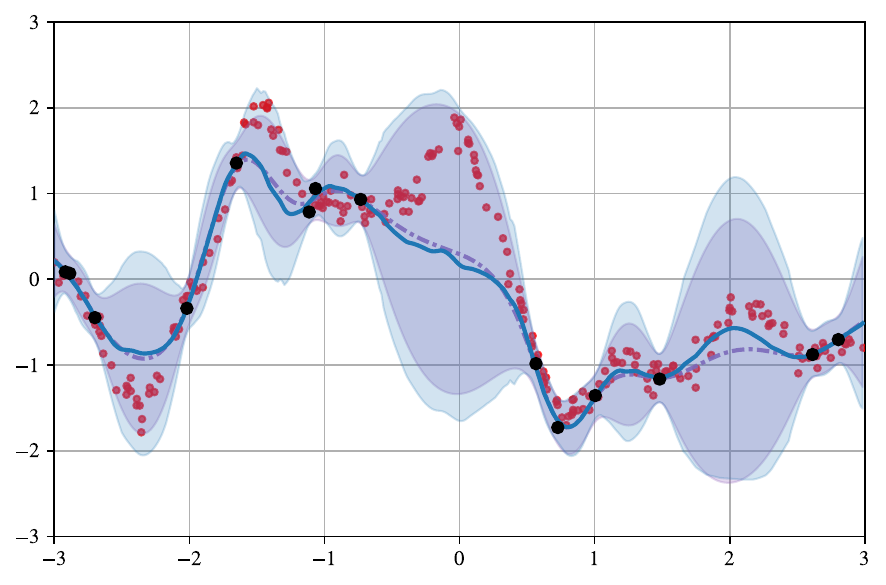} & 
            \includegraphics[width=0.33\textwidth]{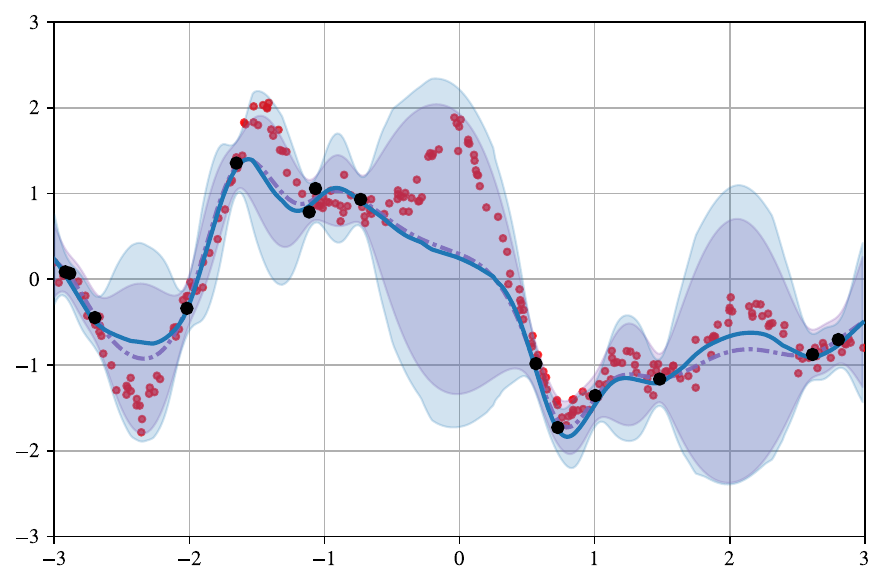} &
            \includegraphics[width=0.33\textwidth]{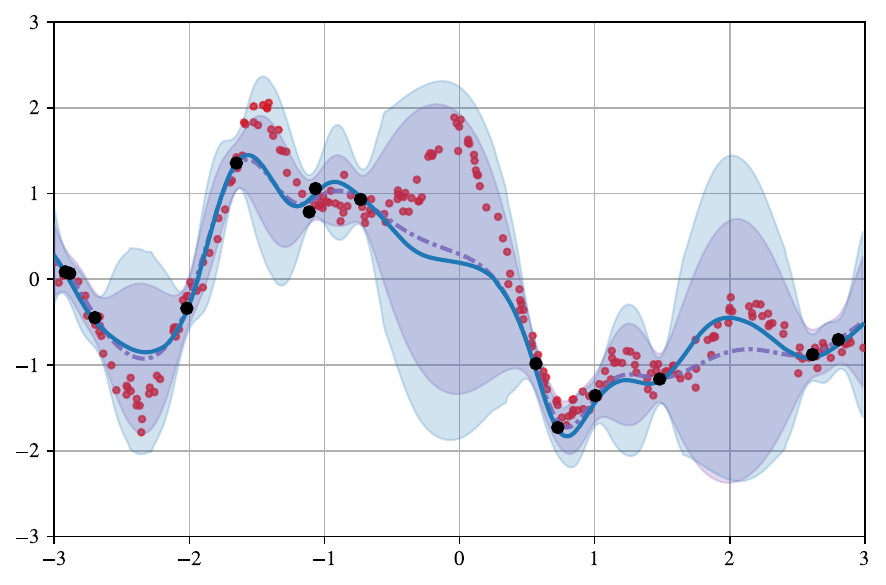}
            \\
            \raisebox{3.85\normalbaselineskip}[0pt][0pt]{\rotatebox[origin=c]{90}{\scriptsize GP-Periodic}} &
            \includegraphics[width=0.33\textwidth]{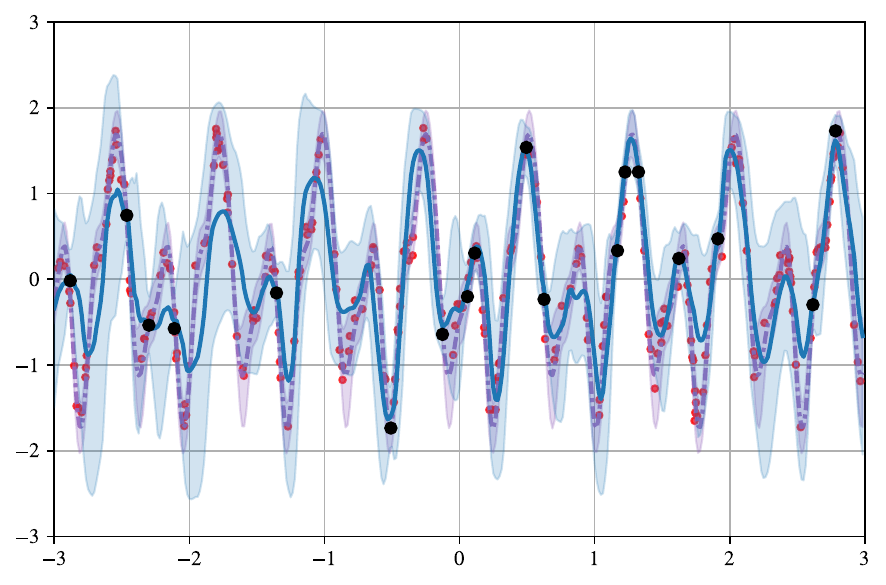} & 
            \includegraphics[width=0.33\textwidth]{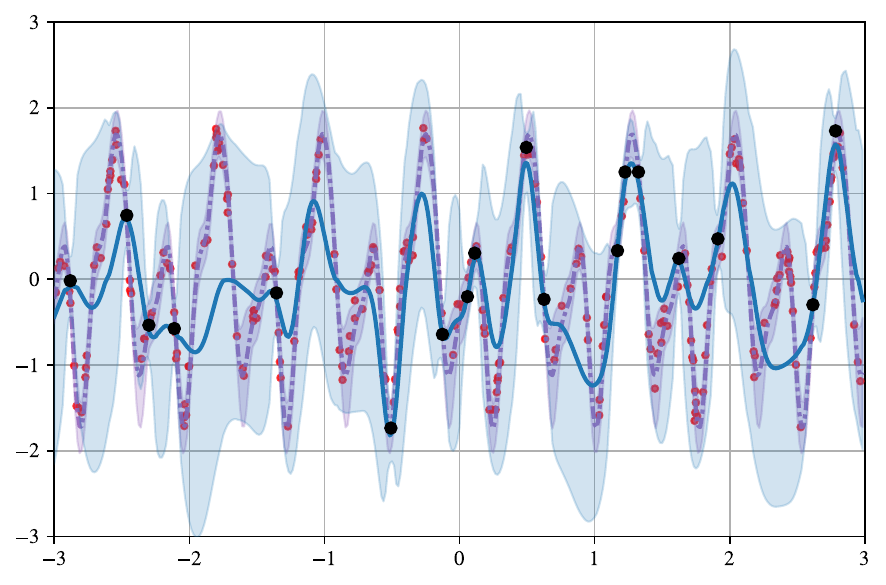} &
            \includegraphics[width=0.33\textwidth]{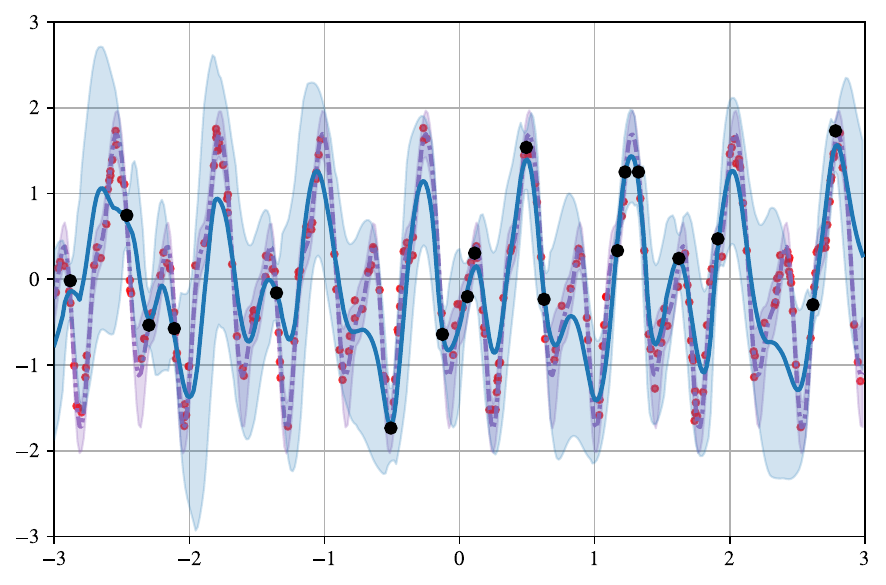}
            \\
            \raisebox{2.5\normalbaselineskip}[0pt][0pt]{\rotatebox[origin=c]{90}{\scriptsize Sawtooth}} &
            \includegraphics[width=0.33\textwidth]{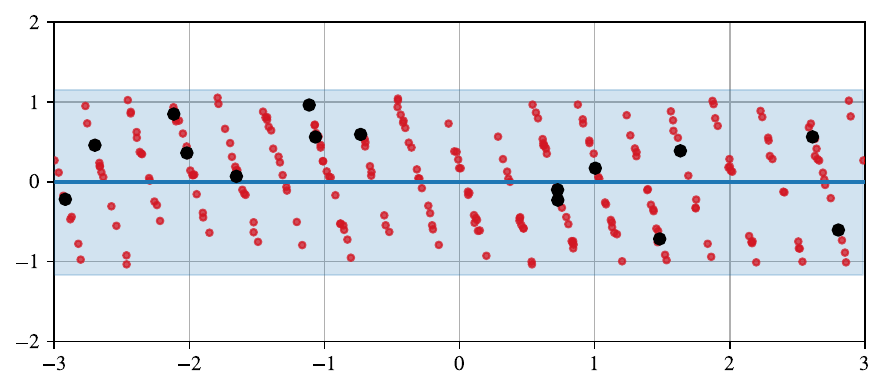} & 
            \includegraphics[width=0.33\textwidth]{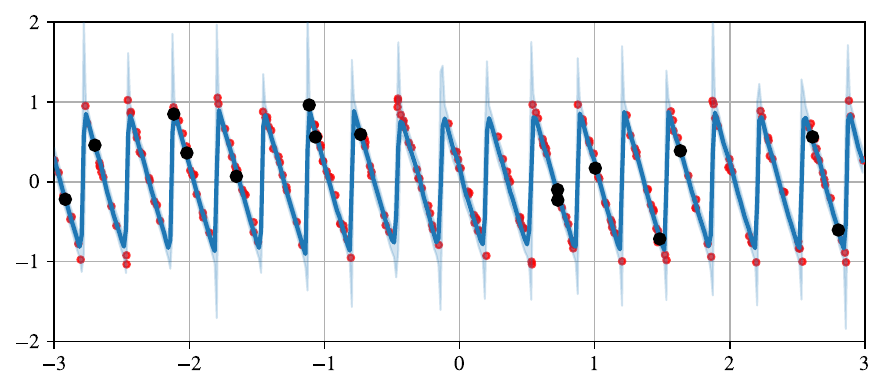} &
            \includegraphics[width=0.33\textwidth]{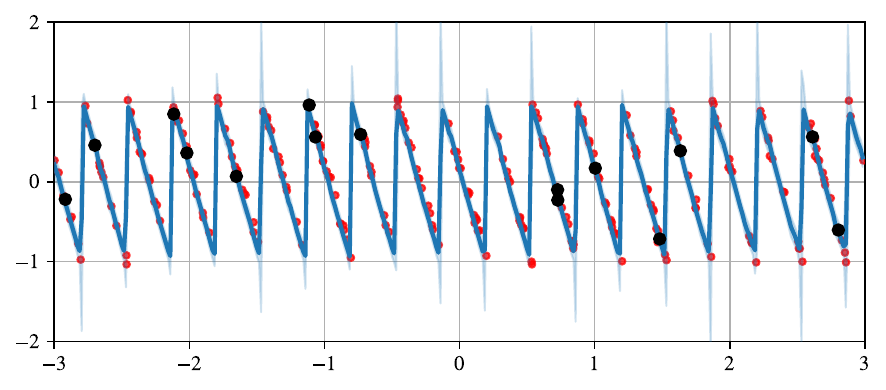}
            \\
            \raisebox{2.75\normalbaselineskip}[0pt][0pt]{\rotatebox[origin=c]{90}{\scriptsize Square}} &
            \includegraphics[width=0.33\textwidth]{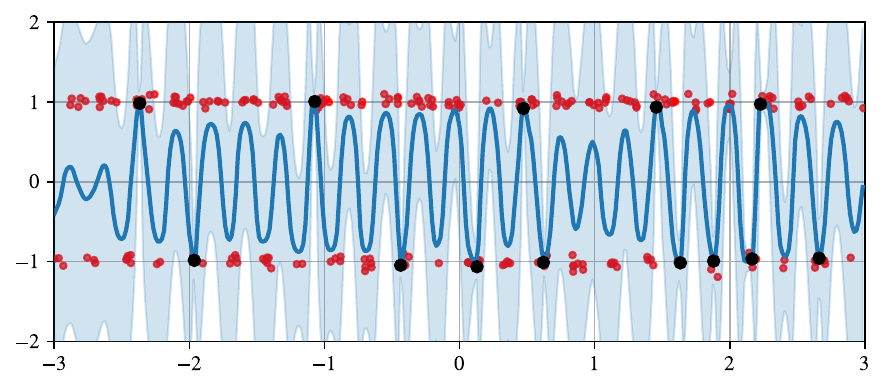} & 
            \includegraphics[width=0.33\textwidth]{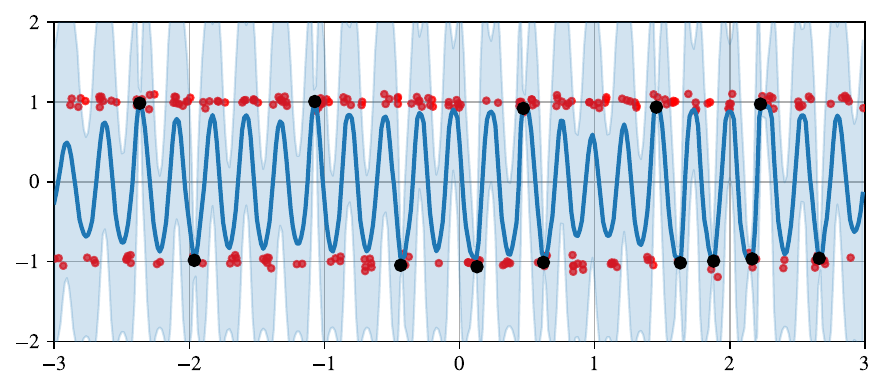} &
            \includegraphics[width=0.33\textwidth]{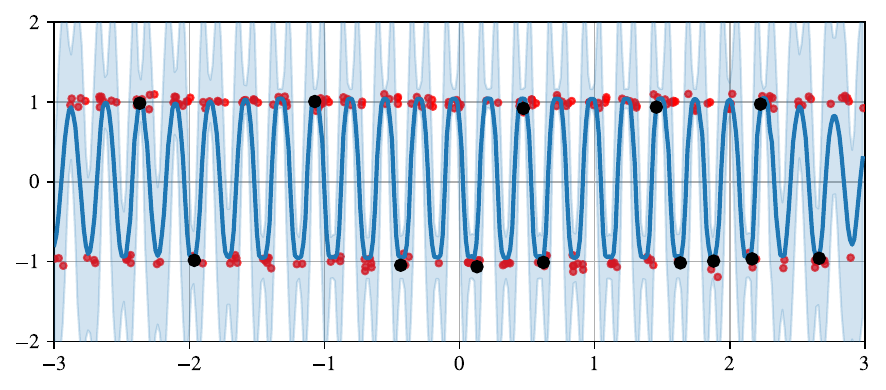}
            \\
             &TE-TNP & ConvCNP & SConvCNP
        \end{tabular}
        \endgroup
    }
    \caption{Example predictions on synthetic data. For each model, the blue curve denotes the predictive mean, and the shaded region corresponds to $\pm 2$ standard deviations of the model’s Gaussian predictive distribution. In the first two rows, where the data are generated from Gaussian processes, the ground-truth distribution is shown in purple: the dash--dotted curve indicates the true mean, and the shaded band represents $\pm 2$ standard deviations around it. Black points denote context observations, and the red point indicates the queries.}
    \label{fig: 1d-synthetic}
    \vspace{-0.25cm}
\end{figure*}
We evaluate our framework on four regression benchmarks and compare its performance against several members of CNPs family. Specifically, we include the original CNP~\citep{garnelo2018conditional}, the Attentive CNP (AttCNP; \citet{kim2019attentive}), the Convolutional CNP (ConvCNP; \citet{gordon2019convolutional}), the diagonal variant of the Transformer Neural Process (TNP; \citet{nguyen2022transformer}), and the Translation-Equivariant Transformer Neural Process (TE-TNP; \citet{ashman2024translation}). For each experimental setting, all models are trained using four random seeds. Performance is evaluated using the final aggregated log-likelihood and root-mean-squared error (RMSE), and we report the mean $\pm$ standard deviation across runs. Additional details regarding datasets, architectures, and training procedures are provided in Appendix~\ref{appendix: experimental-details}. Our implementation and experimental code are publicly available at \href{https://github.com/peiman-m/SConvCNP}{https://github.com/peiman-m/SConvCNP}.

\subsection{Synthetic 1-D Regression}\label{subsec: synthetic-1d}
\begin{table*}[!t]\centering
    \caption{Comparison of predictive performance across methods on synthetically generated tasks. Lower RMSE and higher log-likelihood indicate better performance. For each metric and experimental setting, boldface denotes the top-two performing models.}
    \vspace{2.5pt}
    \label{table: 1d-synthetic-experiment-results}
    \renewcommand{\arraystretch}{1.3}
    \scalebox{0.9}{
        \begin{tabular}{@{}l@{\hskip 7.5pt} l@{\hskip 5pt} c@{\hskip 2.5pt} c@{\hskip 2.5pt} c@{\hskip 2.5pt} c@{\hskip 2.5pt} c@{\hskip 2.5pt} c@{\hskip 1pt}}
        \toprule
        \multirow{2}{*}{Metric} & \multirow{2}{*}{Data} & \multicolumn{6}{c}{Model} \\
        \cmidrule(lr){3-8}
        & & {\footnotesize CNP} & {\footnotesize AttCNP} & {\footnotesize TNP} & {\footnotesize TE-TNP} & {\footnotesize ConvCNP} & {\footnotesize SConvCNP} \\
        \midrule
        \multirow{4}{*}{\footnotesize Log-likelihood ↑} 
        & {\footnotesize Mat\'ern 5/2} & ${-0.54_{\pm0.01}}$ & ${-0.32_{\pm0.00}}$ & ${\bm{-0.29}_{\pm0.00}}$ & ${\bm{-0.28}_{\pm0.00}}$ & ${-0.30_{\pm0.01}}$ & ${\bm{-0.29}_{\pm0.01}}$ \\
        & {\footnotesize Periodic} & ${-1.20_{\pm0.00}}$ & ${-0.87_{\pm0.05}}$ & ${-0.76_{\pm0.03}}$ & ${\bm{-0.57}_{\pm0.02}}$ & ${-0.73_{\pm0.01}}$ & ${\bm{-0.66}_{\pm0.00}}$ \\
        & {\footnotesize Sawtooth} & ${-0.90_{\pm0.00}}$ & ${-0.90_{\pm0.00}}$ & ${-0.90_{\pm0.00}}$ & ${-0.90_{\pm0.00}}$ & ${\bm{0.10}_{\pm0.30}}$ & ${\bm{0.82}_{\pm0.03}}$ \\
        & {\footnotesize Square} & ${-1.39_{\pm0.00}}$ & ${-1.41_{\pm0.01}}$ & ${-1.33_{\pm0.02}}$ & ${\bm{-1.17}_{\pm0.06}}$ & ${-1.35_{\pm0.04}}$ & ${\bm{-1.13}_{\pm0.03}}$ \\
        \midrule
        \multirow{4}{*}{\footnotesize RMSE ↓} 
        & {\footnotesize Mat\'ern 5/2} & ${0.50_{\pm0.00}}$ & ${\bm{0.45}_{\pm0.00}}$ & ${\bm{0.44}_{\pm0.00}}$ & ${\bm{0.44}_{\pm0.00}}$ & ${\bm{0.45}_{\pm0.00}}$ & ${\bm{0.45}_{\pm0.00}}$ \\
        & {\footnotesize Periodic} & ${0.81_{\pm0.00}}$ & ${0.65_{\pm0.02}}$ & ${0.60_{\pm0.02}}$ & ${\bm{0.50}_{\pm0.01}}$ & ${0.57_{\pm0.00}}$ & ${\bm{0.53}_{\pm0.00}}$ \\
        & {\footnotesize Sawtooth} & ${0.58_{\pm0.00}}$ & ${0.58_{\pm0.00}}$ & ${0.58_{\pm0.00}}$ & ${0.58_{\pm0.00}}$ & ${\bm{0.40}_{\pm0.06}}$ & ${\bm{0.24}_{\pm0.00}}$ \\
        & {\footnotesize Square} & ${0.98_{\pm0.00}}$ & ${1.00_{\pm0.01}}$ & ${0.93_{\pm0.02}}$ & ${\bm{0.81}_{\pm0.01}}$ & ${0.89_{\pm0.02}}$ & ${\bm{0.79}_{\pm0.01}}$ \\
    \bottomrule
\end{tabular}}
\end{table*}
We begin by evaluating models on four synthetic benchmarks generated from distinct stochastic processes: a GP with a Matérn--5/2 kernel, a GP with a periodic kernel, a sawtooth-wave generator, and a square-wave generator. For each benchmark, the parameters of the generative process---kernel hyperparameters for the GPs, frequency and direction for the sawtooth generator, and frequency and duty cycle for the square-wave generator---are sampled randomly. Complete experimental details are provided in Appendix~\ref{appendix: synthetic-regression-experiment-details}. 
Table~\ref{table: 1d-synthetic-experiment-results} reports the average evaluation metrics over 1{,}000 test batches, each containing 16 tasks (each task defined in \eqref{eq: task}). For every task—independent of batch—both $x_{c}$ and $x_{q}$ are sampled independently from $\mathcal{U}[-3, 3)$. For each test batch, the number of context points, shared by all tasks within that batch, is drawn independently according to $n_c \sim \mathcal{U}[5, 25)$. The number of query points, however, is fixed at $n_q = 256$ for all the test tasks.
As shown, SConvCNP consistently outperforms or closely matches strong baselines, including TE-TNP and ConvCNP, which represent the current state of the art. Figure~\ref{fig: 1d-synthetic} provides qualitative comparisons of predictive maps produced by SConvCNP, ConvCNP, and TE-TNP.

For the sawtooth-wave benchmark, we were unable to successfully train either TNP or TE-TNP. Across the configurations we attempted—including increased model capacity and multiple randomized initializations—their predictions consistently collapsed to zero. We hypothesize that this failure mode stems from \emph{spectral bias}, the tendency of neural networks to favor low-frequency structure over high-frequency components \citep{rahaman2019spectral, ronen2019convergence, basri2020frequency, tancik2020fourier, fridovich2022spectral}. Recent findings by \citet{vasudeva2025transformers} further suggest that this bias is \emph{exacerbated} in transformer architectures compared with convolutional networks. This interpretation aligns with observations by \citet{nguyen2022transformer}, who originally introduced TNPs. They report degraded performance on GP samples drawn from a periodic kernel—reflected in poor log-likelihood scores—despite strong results on Matérn-kernel tasks. Interestingly, our experiments on periodic-kernel GP tasks (Table~\ref{table: 1d-synthetic-experiment-results}) do \emph{not} replicate this limitation. A key distinction is that we employ the Efficient Query TNP (EQTNP) architecture of \citet{feng2022latent}, rather than the original TNP design \citep{nguyen2022transformer} (see Appendix~\ref{appendix: synthetic-regression-model-architectures} for architectural details). Finally, although sawtooth and square-wave functions share similar discontinuity and high-frequency characteristics, Table~\ref{table: 1d-synthetic-experiment-results} shows that TNP and TE-TNP do \emph{not} collapse when trained on square-wave signals. Investigating the source of this discrepancy falls beyond the scope of this work but represents an intriguing direction for future study.

\subsection{Predator–Prey Dynamics}\label{subsec: pred-prey-subsec}
We next assess performance on simulated trajectories from a stochastic variant of the Lotka--Volterra predator--prey system \citep{Lotka1910contribution, Volterra1926variazioni}, following the formulation in \citet{bruinsma2023autoregressive}. Let $U_t$ and $V_t$ denote the prey and predator populations at time $t$, respectively. Their dynamics evolve according to the stochastic Lotka--Volterra system
\begin{equation*}
    dU_t = \alpha U_t\,dt - \beta U_t V_t\,dt 
           + \sigma U^{\nu}_t\, dB^{(1)}_t, \qquad
    dV_t = -\gamma U_t\,dt + \delta U_t V_t\,dt 
           + \sigma V^{\nu}_t\, dB^{(2)}_t ,
\end{equation*}
where $B^{(1)}_t$ and $B^{(2)}_t$ are independent Brownian motions.In the deterministic component of the dynamics, $U_t$ grows exponentially at rate $\alpha$, while $V_t$ decays at rate $\gamma$. The bilinear interaction terms $\beta U_t V_t$ and $\delta U_t V_t$ model predation and the corresponding transfer of biomass from prey to predators. To account for stochastic fluctuations commonly observed in empirical population counts, the dynamics are augmented with multiplicative noise terms $\sigma U^{\nu}_t dB^{(1)}_t$ and $\sigma V^{\nu}_t dB^{(2)}_t$. Here, $\sigma$ controls the noise magnitude, while $\nu$ determines how the variability scales with population size.

We simulate trajectories over a dense uniform grid on $t \in [-10, 100)$, discarding the initial 10 years as burn-in. For each task, context points $x_{c}$ and query points $x_{q}$ are drawn independently and uniformly from the retained interval $[0,100)$. Population values (corresponding to $y = [U_t, V_t]$) and time inputs (corresponding to $x = t$) are rescaled by factors of $0.01$ and $0.1$, respectively, before being fed to the models. A complete description of the experimental protocol is provided in Appendix~\ref{appendix: pred–prey-experiment-details}.

Table~\ref{table: pred-prey-experiment-results} reports average evaluation metrics over 1{,}000 test batches, each containing 16 tasks. For each batch, the number of context points—shared across all tasks—is sampled as $n_c \sim \mathcal{U}[5,25)$, and the number of query points is fixed at $n_q = 256$. SConvCNP achieves log-likelihood performance comparable to ConvCNP, and both are competitive with TE-TNP. In terms of RMSE, SConvCNP matches TE-TNP, indicating that its predictive uncertainty intervals are slightly wider. Qualitative comparisons of predictive maps for the top three models appear in Figure~\ref{fig: pred-prey}.

\begin{table*}[!t]\centering
    \caption{
    Comparison of predictive performance across methods on tasks constructed from the Lotka--Volterra system simulation. Lower RMSE and higher log-likelihood indicate better performance. For each metric and experimental setting, boldface denotes the top-two performing models.}
    \vspace{2.5pt}
    \renewcommand{\arraystretch}{1.3}
    \label{table: pred-prey-experiment-results}
    \scalebox{0.9}{
    \begin{tabular}{l@{\hskip 15pt} c@{\hskip 10pt} c@{\hskip 10pt} c@{\hskip 10pt} c@{\hskip 10pt} c@{\hskip 10pt} c@{\hskip 1pt}}
        \toprule
        & {\footnotesize CNP} & {\footnotesize AttCNP} & {\footnotesize TNP} & {\footnotesize TE-TNP} & {\footnotesize ConvCNP} & {\footnotesize SConvCNP} \\
        \midrule
        {\footnotesize Log-likelihood ↑}   & ${-0.27_{\pm0.01}}$ & ${0.08_{\pm0.00}}$  & ${0.12_{\pm0.00}}$ & ${\bm{0.16}_{\pm0.00}}$ & ${\bm{0.14}_{\pm0.00}}$ & ${\bm{0.14}_{\pm0.00}}$ \\
        {\footnotesize RMSE ↓}     & ${0.37_{\pm0.00}}$ & ${\bm{0.31}_{\pm0.00}}$  & ${\bm{0.31}_{\pm0.00}}$ & ${\bm{0.30}_{\pm0.0.00}}$ & ${\bm{0.31}_{\pm0.00}}$ & ${\bm{0.30}_{\pm0.00}}$ \\
        \bottomrule
    \end{tabular}}
\end{table*}

\begin{figure*}[!t]
    \vspace{-0.1cm}
    \hspace*{-0.35cm}
    \centering
    \scalebox{1.0}{
        \begingroup
        \setlength{\tabcolsep}{0pt}
        \renewcommand{\arraystretch}{1.0}
        \begin{tabular}{l@{\hskip -0.5pt}ccc}
            &&
            \makebox[0pt][c]{%
                    \includegraphics[width=0.4\textwidth]{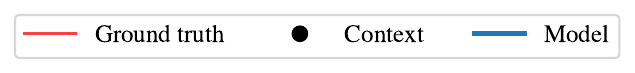}%
            }
            &\\
            \raisebox{3.85\normalbaselineskip}[0pt][0pt]{\rotatebox[origin=c]{90}{\scriptsize  \hspace{7pt} Predator \hspace{25pt} Prey}} &
            \includegraphics[width=0.33\textwidth]{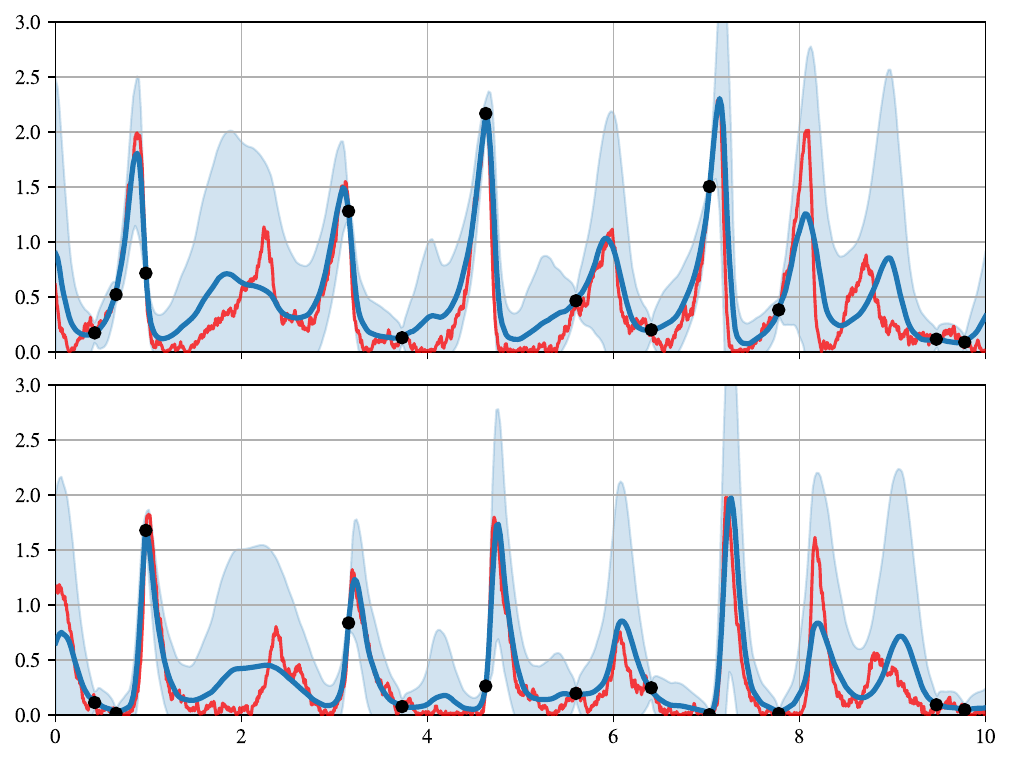} & 
            \includegraphics[width=0.33\textwidth]{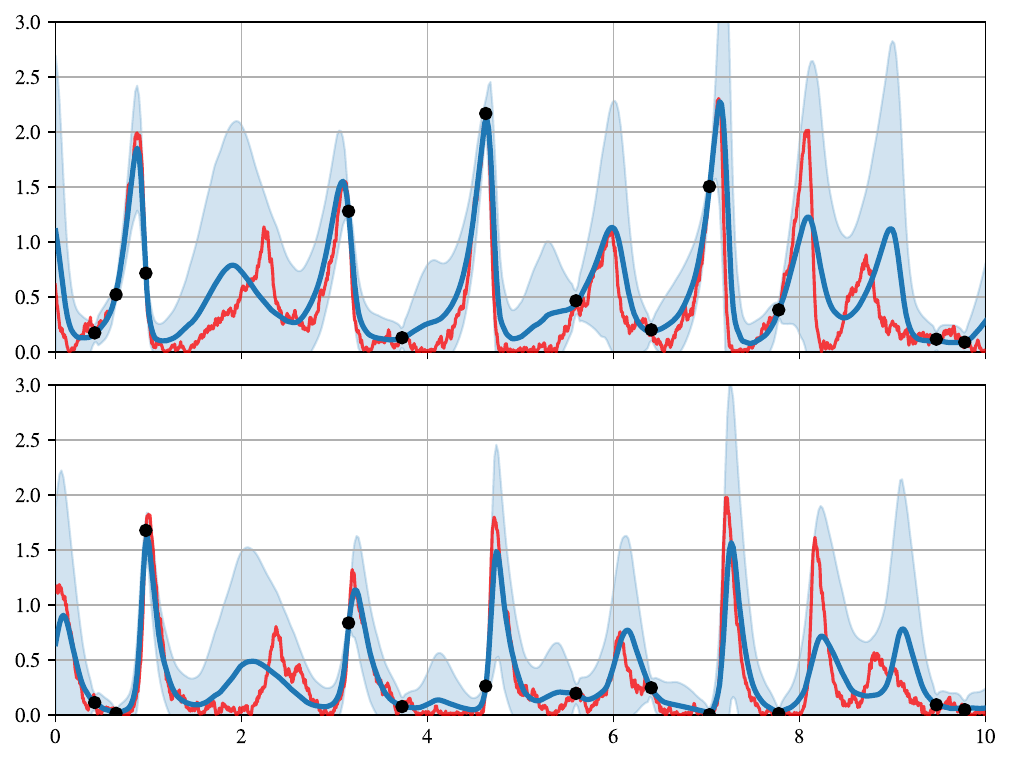} &
            \includegraphics[width=0.33\textwidth]{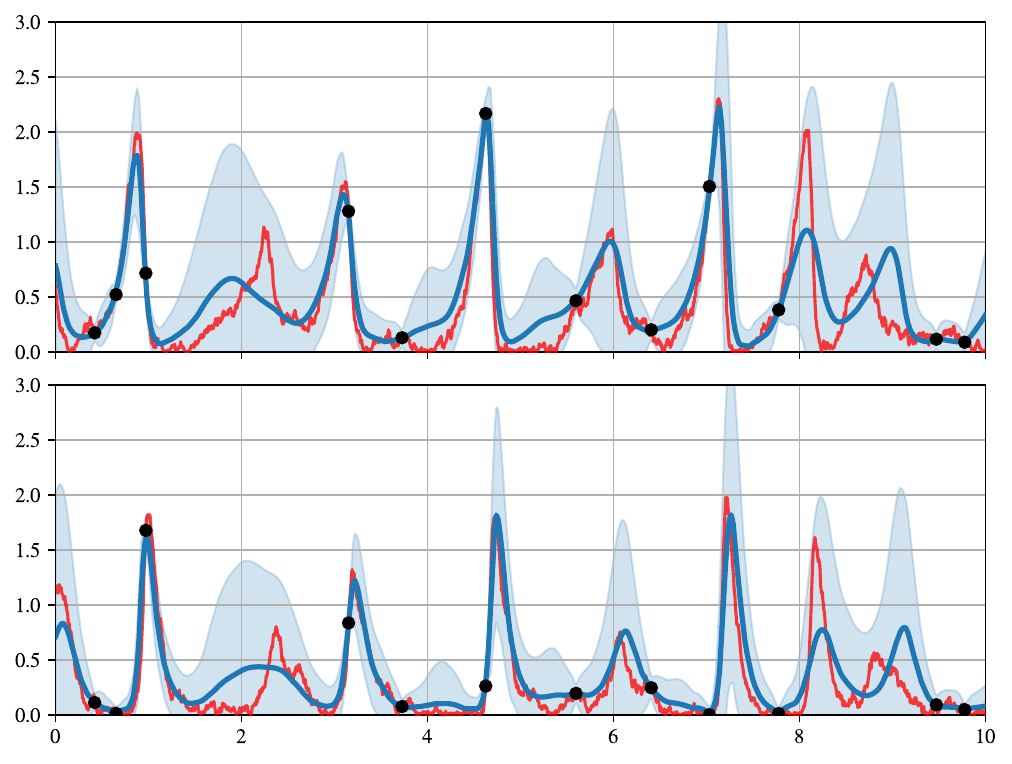}\\
            &TE-TNP & ConvCNP & SConvCNP \\
        \end{tabular}
        \endgroup
    }
    \caption{Illustrative predictions on simulated Lotka--Volterra predator--prey trajectories. Black markers denote the context observations. The blue curve shows the predictive mean, while the shaded region corresponds to the $\pm 2$ standard deviation interval of the Gaussian predictive distribution. Ground-truth population trajectories from the simulator are plotted in red.}
    \label{fig: pred-prey}
    \vspace{-0.25cm}
\end{figure*}

\subsection{Traffic Flow}\label{subsection: traffic-flow-experiment}
\begin{table*}[!b]\centering
    \vspace{-0.5cm}
    \caption{Comparison of predictive performance across methods on  tasks constructed from California traffic flow measurements. Lower RMSE and higher log-likelihood indicate better performance. For each metric and experimental setting, boldface denotes the top-two performing models.}
    \vspace{2.5pt}
    \renewcommand{\arraystretch}{1.3}
    \label{table: traffic-flow-experiment-results}
    \scalebox{0.9}{
    \begin{tabular}{l@{\hskip 15pt} c@{\hskip 10pt} c@{\hskip 10pt} c@{\hskip 10pt} c@{\hskip 10pt} c@{\hskip 10pt} c@{\hskip 1pt}}
        \toprule
        & {\footnotesize CNP} & {\footnotesize AttCNP} & {\footnotesize TNP} & {\footnotesize TE-TNP} & {\footnotesize ConvCNP} & {\footnotesize SConvCNP} \\
        \midrule
        {\footnotesize Log-likelihood ↑}   & ${1.73_{\pm0.10}}$ & ${1.80_{\pm0.01}}$  & ${1.93_{\pm0.06}}$ & ${1.72_{\pm0.02}}$ & ${\bm{1.98}_{\pm0.02}}$ & ${\bm{2.05}_{\pm0.02}}$ \\
        {\footnotesize RMSE ↓}     & ${\bm{0.05}_{\pm0.00}}$ & ${\bm{0.05}_{\pm0.00}}$  & ${\bm{0.04}_{\pm0.00}}$ & ${\bm{0.05}_{\pm0.00}}$ & ${\bm{0.04}_{\pm0.00}}$ & ${\bm{0.04}_{\pm0.00}}$ \\
        \bottomrule
    \end{tabular}}
\end{table*}
For our third experiment, we evaluate on the California traffic-flow dataset from LargeST~\citep{liu2023largest}. This dataset comprises five years (2017--2021) of traffic measurements recorded every 5 minutes by approximately 8{,}600 loop-detector sensors deployed across California's highway network. We focus on the year 2020, where traffic patterns are expected to exhibit heightened variability due to the abrupt onset of the COVID-19 pandemic. For each sensor, we segment its year-long time series into non-overlapping 14-day windows and downsample each window by a factor of 6 (from 5-minute to 30-minute resolution). Each window is treated as a dense trajectory from which individual tasks are constructed. Additional details of the experimental setup are provided in Appendix~\ref{appendix: traffic-flow-experiment-details}. 

\begin{figure*}[!t]
    \vspace{-0.2cm}
    \hspace*{-0.25cm}
    \centering
    \scalebox{1.0}{
        \begingroup
        \setlength{\tabcolsep}{0pt}
        \renewcommand{\arraystretch}{1.0}
        \begin{tabular}{ccc}
            &
            \makebox[0pt][c]{%
                    \includegraphics[width=0.4\textwidth]{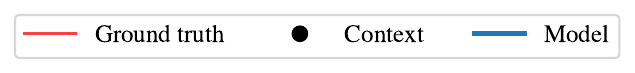}%
            }
            &\\
            \includegraphics[width=0.33\textwidth]{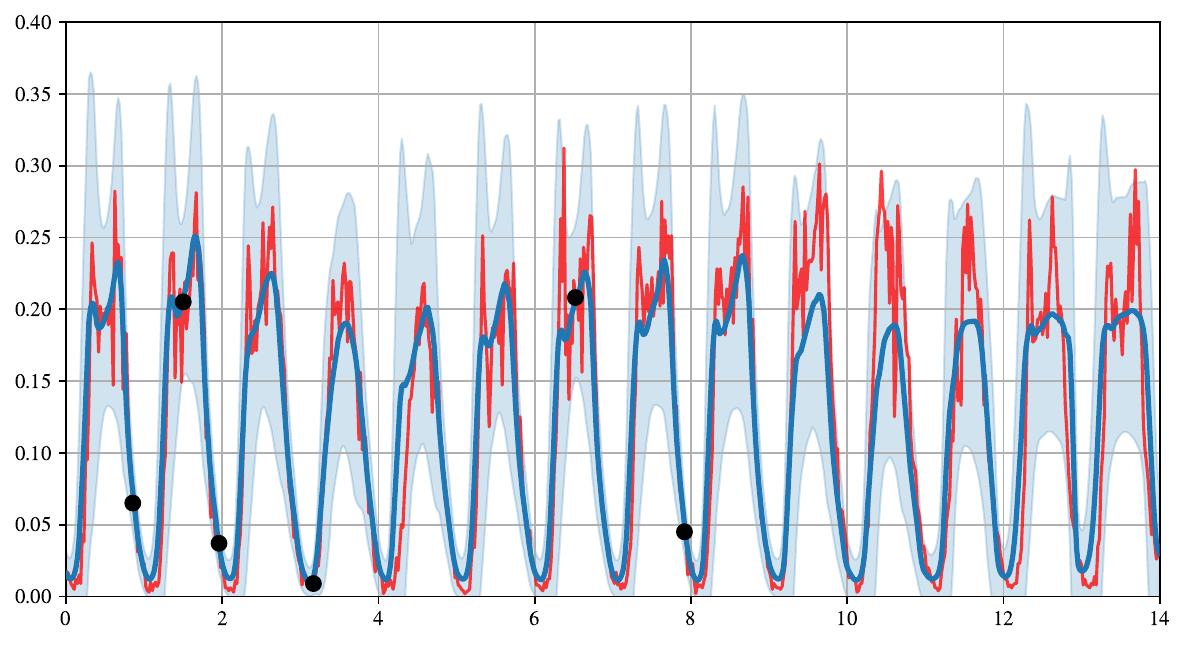} & 
            \includegraphics[width=0.33\textwidth]{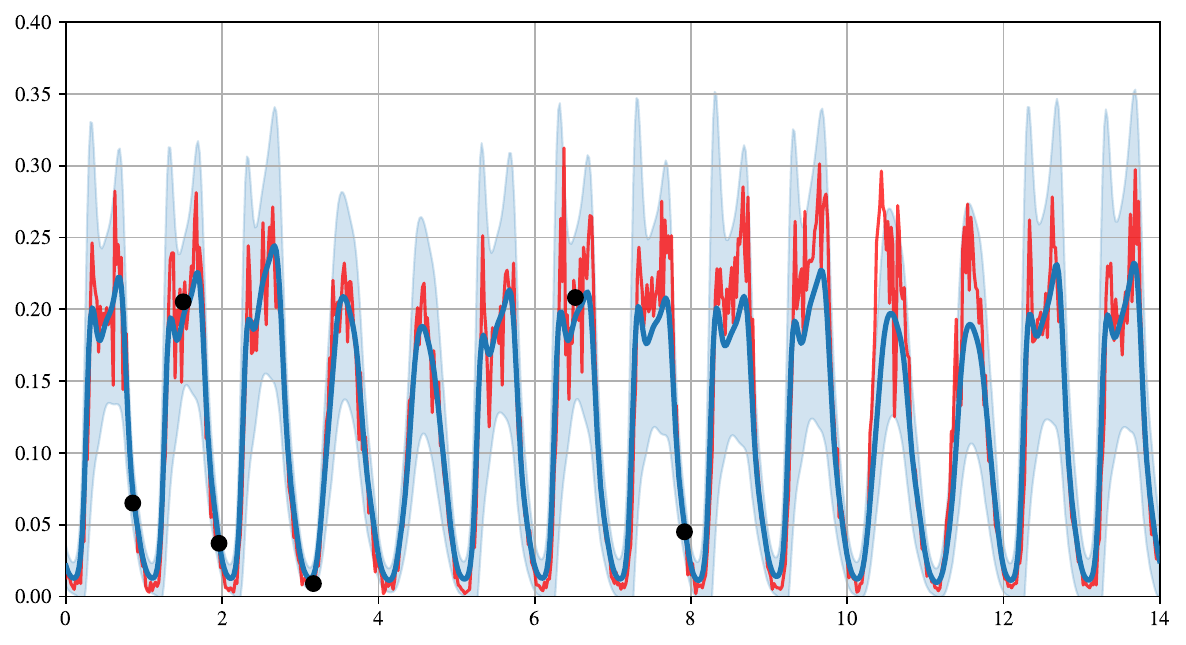} &
            \includegraphics[width=0.33\textwidth]{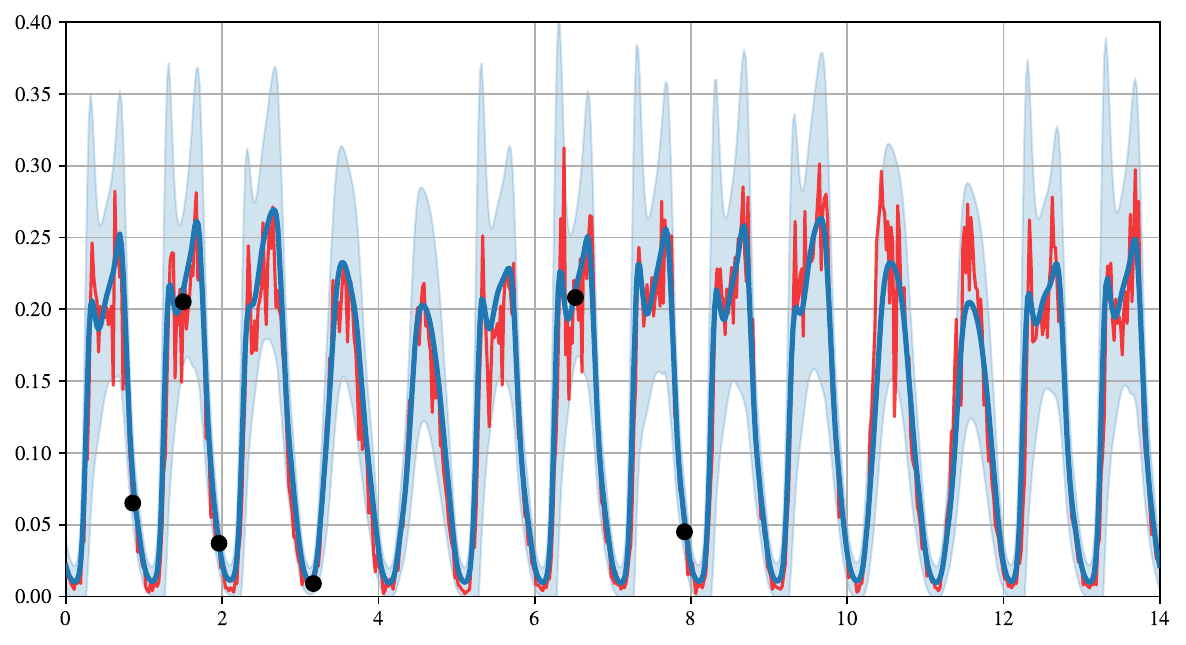}\\
            TE-TNP & ConvCNP & SConvCNP \\
        \end{tabular}
        \endgroup
    }
    \caption{Illustrative predictions on the California traffic-flow dataset. Black markers denote the context observations. The blue curve shows the predictive mean, while the shaded region corresponds to the $\pm 2$ standard deviation interval of the Gaussian predictive distribution. Ground-truth traffic measurements are shown in red.}
    \label{fig: traffic-flow}
    \vspace{-0.25cm}
\end{figure*}

Table~\ref{table: traffic-flow-experiment-results} reports evaluation metrics averaged over 66{,}586 test tasks, constructed from 2{,}561 held-out test sensors and partitioned into batches of size 32. For each batch, the number of context points---shared across all tasks---is drawn as $n_c \sim \mathcal{U}[5, 25)$, and the number of query points is fixed at $n_q = 50$. Across all metrics, SConvCNP achieves the best performance, attaining the highest log-likelihood and lowest RMSE among the baselines. Figure~\ref{fig: traffic-flow} provides qualitative comparisons of the predictive maps produced by the three models.

\subsection{Image Completion}\label{subsection: image-completion-experiment}
\begin{figure*}[h]
\centering
\begin{minipage}[h]{0.46\textwidth}
    For our final experiment, we evaluate model performance on an image-completion task formulated as a spatial regression problem, where the model maps 2D pixel coordinates to their corresponding intensity values. We use images from the Describable Textures Dataset (DTD; \citet{cimpoi2014describing}), and construct each task from a processed $64 \times 64$ subsampled crop of an original image. Because each task contains a large number of context and query pixels, we were unable to fit TE-TNP within our computational budget—even after reducing its size—so we exclude it from this experiment. Additional experimental details are provided in Appendix~\ref{appendix: image-completion-experiment-details}.
    Table~\ref{table: image-completion-experiment-results} reports results averaged over 1{,}880 test tasks, evaluated in batches of 16. For each batch, the number of context points—shared across all tasks—is sampled as $n_c \sim \mathcal{U}[5, 1024)$; all remaining pixels serve as query points, so that $n_q = 4096 - n_c$. Among the baselines, SConvCNP achieves the highest log-likelihood and the second-lowest RMSE. Figure~\ref{fig: image-completion} presents qualitative comparisons of the predictive means produced by the top three models.
\end{minipage}
\hfill
\begin{minipage}[h]{0.5\textwidth}
    \vspace{-0.5cm}
    \centering
    \scalebox{1.0}{
        \begingroup
        \setlength{\tabcolsep}{0pt}
        \begin{tabular}{ccccc}
            \includegraphics[width=0.2\textwidth]{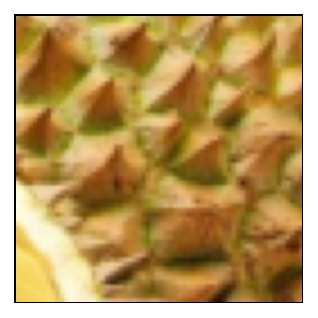} &
            \includegraphics[width=0.2\textwidth]{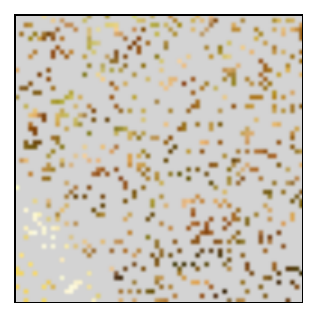} &
            \includegraphics[width=0.2\textwidth]{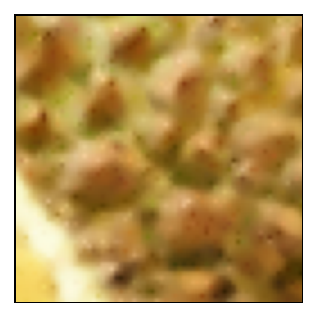} &
            \includegraphics[width=0.2\textwidth]{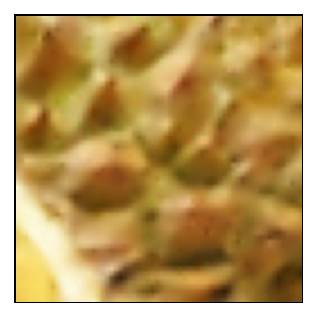} &
            \includegraphics[width=0.2\textwidth]{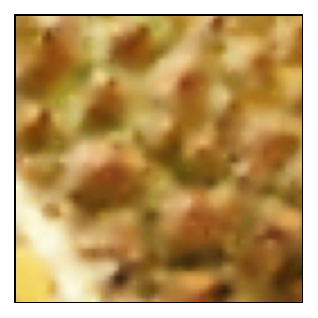} \\
            \includegraphics[width=0.2\textwidth]{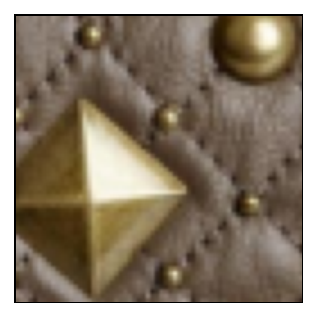} &
            \includegraphics[width=0.2\textwidth]{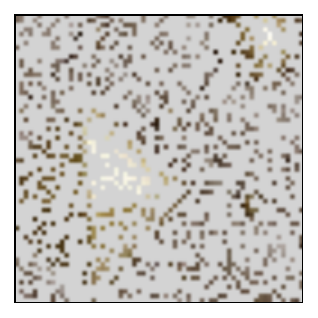} &
            \includegraphics[width=0.2\textwidth]{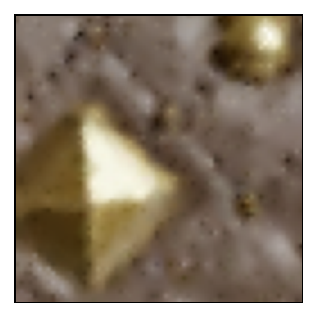} &
            \includegraphics[width=0.2\textwidth]{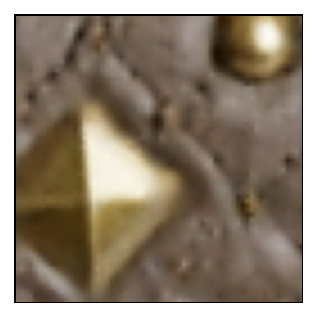} &
            \includegraphics[width=0.2\textwidth]{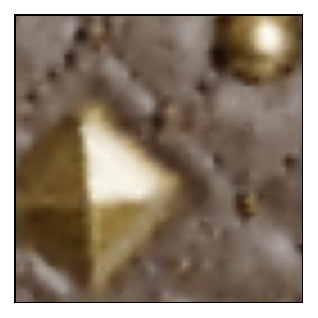} \\
            \includegraphics[width=0.2\textwidth]{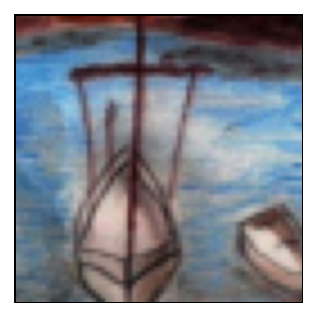} &
            \includegraphics[width=0.2\textwidth]{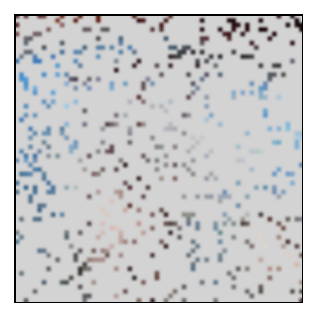} &
            \includegraphics[width=0.2\textwidth]{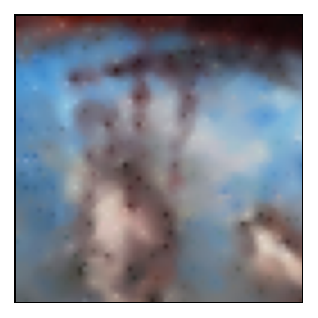} &
            \includegraphics[width=0.2\textwidth]{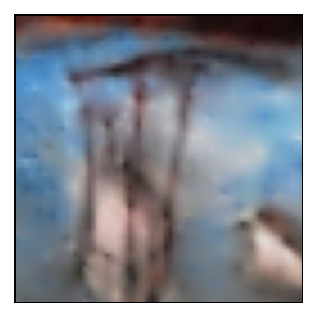} &
            \includegraphics[width=0.2\textwidth]{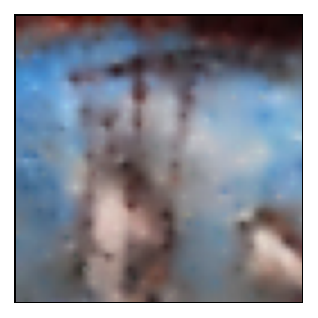} \\
            \includegraphics[width=0.2\textwidth]{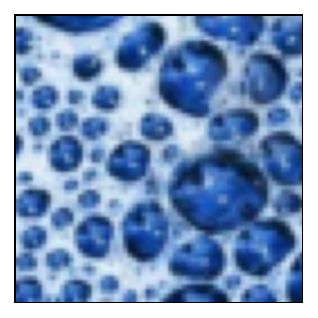} &
            \includegraphics[width=0.2\textwidth]{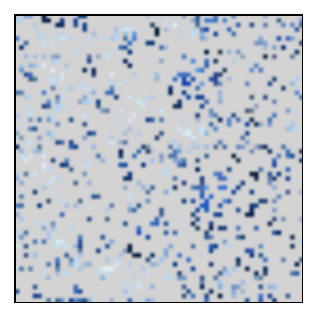} &
            \includegraphics[width=0.2\textwidth]{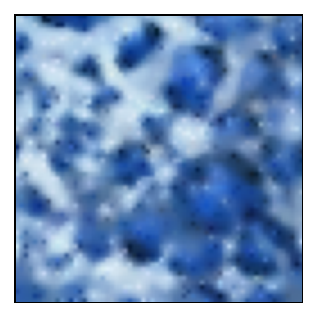} &
            \includegraphics[width=0.2\textwidth]{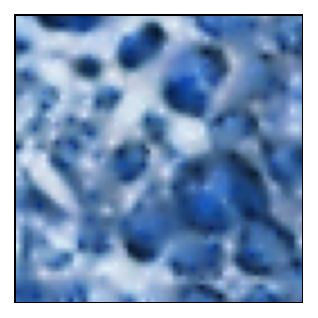} &
            \includegraphics[width=0.2\textwidth]{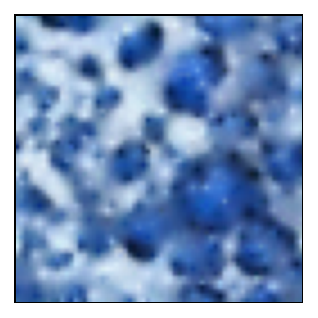} \\
            {\scriptsize Ground Truth} &{\scriptsize Context} &{\scriptsize TNP} & {\scriptsize ConvCNP} & {\scriptsize SConvCNP}
        \end{tabular}
        \endgroup
    }
    \caption{Illustrative model outputs for image completion on the DTD dataset. Gray pixels indicate query regions, while the remaining pixels serve as context observations. For each model, query regions are completed using the mean of the predictive distribution.}
    \label{fig: image-completion}
\end{minipage}
\end{figure*}

\begin{table*}[!ht]\centering
    \vspace{-0.4cm}
    \caption{
    Comparison of predictive performance across methods on image-completion tasks constructed from the DTD dataset. Lower RMSE and higher log-likelihood indicate better performance. For each metric and experimental setting, boldface denotes the top-two performing models..}
    \vspace{2.5pt}
    \renewcommand{\arraystretch}{1.3}
    \label{table: image-completion-experiment-results}
    \scalebox{0.9}{
    \begin{tabular}{l@{\hskip 15pt} c@{\hskip 10pt} c@{\hskip 10pt} c@{\hskip 10pt} c@{\hskip 10pt} c@{\hskip 1pt}}
        \toprule
        & {\footnotesize CNP} & {\footnotesize AttCNP} & {\footnotesize TNP} & {\footnotesize ConvCNP} & {\footnotesize SConvCNP} \\
        \midrule
        {\footnotesize Log-likelihood ↑}     & ${0.67_{\pm0.01}}$ & ${1.39_{\pm.02}}$  & ${1.39_{\pm0.05}}$ & ${\bm{1.48}_{\pm0.00}}$ & ${\bm{1.50}_{\pm0.02}}$ \\
        {\footnotesize RMSE ↓}   & ${0.14_{\pm0.00}}$ & ${\bm{0.08}_{\pm0.00}}$  & ${\bm{0.06}_{\pm0.00}}$ & ${\bm{0.08}_{\pm0.00}}$ & ${\bm{0.08}_{\pm0.00}}$ \\
        \bottomrule
    \end{tabular}}
\end{table*}

\section{Related Works}\label{sec:related-work}
\paragraph{Neural PDE Solvers}
The high computational cost of traditional numerical PDE solvers has motivated the development of more efficient alternatives based on deep learning \cite{gupta2022towards}. Among these, neural operators \citep{li2020neural, kovachki2023neural}, and in particular Fourier neural operators (FNOs, \citealp{li2020fourier}), have become popular. Numerous extensions have since been proposed: \citet{helwig2023group} introduced rotation- and reflection-equivariant FNOs; \citet{gupta2021multiwavelet} developed multiwavelet neural operators by projecting kernels onto predefined polynomial bases; and \citet{tran2021factorized} reduced model complexity through separable Fourier representations. \citet{xiao2024amortized} proposed the Amortized FNO (AM-FNO), which substantially cuts parameter count by amortizing the Fourier kernel parameterization, while \citet{qin2024toward} analyzed FNOs through the lens of spectral bias. \citet{koshizuka2024understanding} provided a mean-field theoretical perspective, and \citet{zheng2024alias} introduced the Mamba Neural Operator, which couples global context via a state-space model to achieve linear complexity and representation equivalence. \citet{bartolucci2023representation} examined continuous--discrete consistency, showing that it holds for convolutional neural operators \citep{raonic2023convolutional}. Additional works include orthogonal-attention eigenfunction methods for operator learning \citep{xiao2023improved} and hierarchical transformers with frequency-aware priors for resolution-invariant super-resolution \citep{luo2024hierarchical}.

\paragraph{Function Space Inference}
In non-parametric Bayesian modeling, GPs and deep GPs \citep{damianou2013deep} provide flexible function-space priors with well-calibrated uncertainty estimates, but their computational cost becomes prohibitive on large datasets. This challenge has motivated the development of Bayesian neural networks (BNNs; \citet{mackay1992practical, hinton1993keeping, neal2012bayesian}), which combine neural network expressiveness with Bayesian uncertainty quantification. However, specifying meaningful priors over network weights remains notoriously difficult. Recent work therefore reframes Bayesian inference in neural networks as learning a posterior over the \emph{functions} induced by stochastic weights \citep{wolpert1993bayesian, qiu2023should}. Variational implicit processes (VIPs; \citet{ma2019variational, santana2021function, ortega2022deep}) generalize GPs by defining implicit stochastic processes through latent variables, while functional variational BNNs (fBNNs; \cite{sun2019functional}) enforce alignment between BNN-induced priors and target priors by minimizing a functional KL divergence---though this objective can be difficult to compute and, in some cases, ill-posed \citep{burt2020understanding}. Subsequent work has aimed to address these limitations \citep{ma2021functional, rudner2022tractable, wild2022generalized, rudner2023function, wu2024functional}. Orthogonally, energy-based models have been explored for representing stochastic processes in function space \citep{yang2020energy, lim2022energy}. More recently, several approaches have extended diffusion and flow models to function spaces \citep{phillips2022spectral, dutordoir2023neural, mathieu2023geometric, franzese2023continuous, lim2023score, pmlr-v238-kerrigan24a, zhang2025flow}.

\section{Conclusion}
In this work, we introduced spectral convolutional conditional neural processes (SConvCNPs), a class of conditional neural processes that incorporates ideas from operator learning---particularly Fourier neural operators---into the neural processes framework, with an emphasis on convolutional conditional neural processes. Empirical evaluations on a collection of synthetic and real-world datasets indicate that SConvCNPs perform comparably to strong baselines. These results suggest that integrating techniques from operator learning into neural processes is a viable direction for probabilistic function modeling and warrants further investigation.

\acksection
We thank Arman Hasanzadeh and Jonathan W.~Siegel for their insightful feedback and constructive suggestions. We also acknowledge the Texas A\&M High Performance Research Computing facility for providing the computational resources used in this study. Finally, we thank the anonymous reviewers for their valuable comments and suggestions, which helped improve the quality of this work.

\bibliography{bibfile}

@article{gordon2019convolutional,
  title={Convolutional conditional neural processes},
  author={Gordon, Jonathan and Bruinsma, Wessel P and Foong, Andrew YK and Requeima, James and Dubois, Yann and Turner, Richard E},
  journal={arXiv preprint arXiv:1910.13556},
  year={2019}
}

@inproceedings{garnelo2018conditional,
  title={Conditional neural processes},
  author={Garnelo, Marta and Rosenbaum, Dan and Maddison, Christopher and Ramalho, Tiago and Saxton, David and Shanahan, Murray and Teh, Yee Whye and Rezende, Danilo and Eslami, SM Ali},
  booktitle={International conference on machine learning},
  pages={1704--1713},
  year={2018},
  organization={PMLR}
}

@article{garnelo2018neural,
  title={Neural processes},
  author={Garnelo, Marta and Schwarz, Jonathan and Rosenbaum, Dan and Viola, Fabio and Rezende, Danilo J and Eslami, SM and Teh, Yee Whye},
  journal={arXiv preprint arXiv:1807.01622},
  year={2018}
}

@article{li2020fourier,
  title={Fourier neural operator for parametric partial differential equations},
  author={Li, Zongyi and Kovachki, Nikola and Azizzadenesheli, Kamyar and Liu, Burigede and Bhattacharya, Kaushik and Stuart, Andrew and Anandkumar, Anima},
  journal={arXiv preprint arXiv:2010.08895},
  year={2020}
}

@article{rahman2022uno,
  title={U-no: U-shaped neural operators},
  author={Rahman, Md Ashiqur and Ross, Zachary E and Azizzadenesheli, Kamyar},
  journal={arXiv preprint arXiv:2204.11127},
  year={2022}
}

@inproceedings{jung2023bayesian,
  title={Bayesian Convolutional Deep Sets with Task-Dependent Stationary Prior},
  author={Jung, Yohan and Park, Jinkyoo},
  booktitle={International Conference on Artificial Intelligence and Statistics},
  pages={3795--3824},
  year={2023},
  organization={PMLR}
}

@article{kovachki2023neural,
  title={Neural Operator: Learning Maps Between Function Spaces With Applications to PDEs.},
  author={Kovachki, Nikola B and Li, Zongyi and Liu, Burigede and Azizzadenesheli, Kamyar and Bhattacharya, Kaushik and Stuart, Andrew M and Anandkumar, Anima},
  journal={J. Mach. Learn. Res.},
  volume={24},
  number={89},
  pages={1--97},
  year={2023}
}

@article{cooley1965algorithm,
  title={An algorithm for the machine calculation of complex Fourier series},
  author={Cooley, James W and Tukey, John W},
  journal={Mathematics of computation},
  volume={19},
  number={90},
  pages={297--301},
  year={1965}
}

@article{li2020neural,
  title={Neural operator: Graph kernel network for partial differential equations},
  author={Li, Zongyi and Kovachki, Nikola and Azizzadenesheli, Kamyar and Liu, Burigede and Bhattacharya, Kaushik and Stuart, Andrew and Anandkumar, Anima},
  journal={arXiv preprint arXiv:2003.03485},
  year={2020}
}

@article{lecun1998gradient,
  title={Gradient-based learning applied to document recognition},
  author={LeCun, Yann and Bottou, L{\'e}on and Bengio, Yoshua and Haffner, Patrick},
  journal={Proceedings of the IEEE},
  volume={86},
  number={11},
  pages={2278--2324},
  year={1998},
  publisher={Ieee}
}

@article{romero2021ckconv,
  title={Ckconv: Continuous kernel convolution for sequential data},
  author={Romero, David W and Kuzina, Anna and Bekkers, Erik J and Tomczak, Jakub M and Hoogendoorn, Mark},
  journal={arXiv preprint arXiv:2102.02611},
  year={2021}
}

@article{mathieu2021contrastive,
  title={On contrastive representations of stochastic processes},
  author={Mathieu, Emile and Foster, Adam and Teh, Yee},
  journal={Advances in Neural Information Processing Systems},
  volume={34},
  pages={28823--28835},
  year={2021}
}

@inproceedings{dutordoir2023neural,
  title={Neural diffusion processes},
  author={Dutordoir, Vincent and Saul, Alan and Ghahramani, Zoubin and Simpson, Fergus},
  booktitle={International Conference on Machine Learning},
  pages={8990--9012},
  year={2023},
  organization={PMLR}
}

@book{rasmussen2006gaussian,
  title={Gaussian processes for machine learning},
  author={Rasmussen, Carl Edward and Williams, Christopher KI and others},
  volume={1},
  year={2006},
  publisher={Springer}
}

@inproceedings{damianou2013deep,
  title={Deep gaussian processes},
  author={Damianou, Andreas and Lawrence, Neil D},
  booktitle={Artificial intelligence and statistics},
  pages={207--215},
  year={2013},
  organization={PMLR}
}

@inproceedings{hinton1993keeping,
  title={Keeping the neural networks simple by minimizing the description length of the weights},
  author={Hinton, Geoffrey E and Van Camp, Drew},
  booktitle={Proceedings of the sixth annual conference on Computational learning theory},
  pages={5--13},
  year={1993}
}

@book{neal2012bayesian,
  title={Bayesian learning for neural networks},
  author={Neal, Radford M},
  volume={118},
  year={2012},
  publisher={Springer Science \& Business Media}
}

@inproceedings{ma2019variational,
  title={Variational implicit processes},
  author={Ma, Chao and Li, Yingzhen and Hern{\'a}ndez-Lobato, Jos{\'e} Miguel},
  booktitle={International Conference on Machine Learning},
  pages={4222--4233},
  year={2019},
  organization={PMLR}
}

@article{ortega2022deep,
  title={Deep Variational Implicit Processes},
  author={Ortega, Luis A and Santana, Sim{\'o}n Rodr{\'\i}guez and Hern{\'a}ndez-Lobato, Daniel},
  journal={arXiv preprint arXiv:2206.06720},
  year={2022}
}

@article{santana2021function,
  title={Function-space Inference with Sparse Implicit Processes},
  author={Santana, Sim{\'o}n Rodr{\'\i}guez and Zaldivar, Bryan and Hern{\'a}ndez-Lobato, Daniel},
  journal={arXiv preprint arXiv:2110.07618},
  year={2021}
}

@article{sun2019functional,
  title={Functional variational Bayesian neural networks},
  author={Sun, Shengyang and Zhang, Guodong and Shi, Jiaxin and Grosse, Roger},
  journal={arXiv preprint arXiv:1903.05779},
  year={2019}
}

@article{ma2021functional,
  title={Functional variational inference based on stochastic process generators},
  author={Ma, Chao and Hern{\'a}ndez-Lobato, Jos{\'e} Miguel},
  journal={Advances in Neural Information Processing Systems},
  volume={34},
  pages={21795--21807},
  year={2021}
}

@article{rudner2022tractable,
  title={Tractable function-space variational inference in Bayesian neural networks},
  author={Rudner, Tim GJ and Chen, Zonghao and Teh, Yee Whye and Gal, Yarin},
  journal={Advances in Neural Information Processing Systems},
  volume={35},
  pages={22686--22698},
  year={2022}
}

@article{burt2020understanding,
  title={Understanding variational inference in function-space},
  author={Burt, David R and Ober, Sebastian W and Garriga-Alonso, Adri{\`a} and van der Wilk, Mark},
  journal={arXiv preprint arXiv:2011.09421},
  year={2020}
}

@article{wild2022generalized,
  title={Generalized variational inference in function spaces: Gaussian measures meet bayesian deep learning},
  author={Wild, Veit David and Hu, Robert and Sejdinovic, Dino},
  journal={Advances in Neural Information Processing Systems},
  volume={35},
  pages={3716--3730},
  year={2022}
}

@article{kim2019attentive,
  title={Attentive neural processes},
  author={Kim, Hyunjik and Mnih, Andriy and Schwarz, Jonathan and Garnelo, Marta and Eslami, Ali and Rosenbaum, Dan and Vinyals, Oriol and Teh, Yee Whye},
  journal={arXiv preprint arXiv:1901.05761},
  year={2019}
}

@article{kim2022neural,
  title={Neural processes with stochastic attention: Paying more attention to the context dataset},
  author={Kim, Mingyu and Go, Kyeongryeol and Yun, Se-Young},
  journal={arXiv preprint arXiv:2204.05449},
  year={2022}
}

@article{nguyen2022transformer,
  title={Transformer neural processes: Uncertainty-aware meta learning via sequence modeling},
  author={Nguyen, Tung and Grover, Aditya},
  journal={arXiv preprint arXiv:2207.04179},
  year={2022}
}

@article{lee2020bootstrapping,
  title={Bootstrapping neural processes},
  author={Lee, Juho and Lee, Yoonho and Kim, Jungtaek and Yang, Eunho and Hwang, Sung Ju and Teh, Yee Whye},
  journal={Advances in neural information processing systems},
  volume={33},
  pages={6606--6615},
  year={2020}
}

@article{bruinsma2023autoregressive,
  title={Autoregressive conditional neural processes},
  author={Bruinsma, Wessel P and Markou, Stratis and Requiema, James and Foong, Andrew YK and Andersson, Tom R and Vaughan, Anna and Buonomo, Anthony and Hosking, J Scott and Turner, Richard E},
  journal={arXiv preprint arXiv:2303.14468},
  year={2023}
}

@article{bruinsma2021gaussian,
  title={The Gaussian neural process},
  author={Bruinsma, Wessel P and Requeima, James and Foong, Andrew YK and Gordon, Jonathan and Turner, Richard E},
  journal={arXiv preprint arXiv:2101.03606},
  year={2021}
}

@article{markou2022practical,
  title={Practical conditional neural processes via tractable dependent predictions},
  author={Markou, Stratis and Requeima, James and Bruinsma, Wessel P and Vaughan, Anna and Turner, Richard E},
  journal={arXiv preprint arXiv:2203.08775},
  year={2022}
}

@article{foong2020meta,
  title={Meta-learning stationary stochastic process prediction with convolutional neural processes},
  author={Foong, Andrew and Bruinsma, Wessel and Gordon, Jonathan and Dubois, Yann and Requeima, James and Turner, Richard},
  journal={Advances in Neural Information Processing Systems},
  volume={33},
  pages={8284--8295},
  year={2020}
}

@article{lee2023martingale,
  title={Martingale Posterior Neural Processes},
  author={Lee, Hyungi and Yun, Eunggu and Nam, Giung and Fong, Edwin and Lee, Juho},
  journal={arXiv preprint arXiv:2304.09431},
  year={2023}
}

@article{wang2022learning,
  title={Learning expressive meta-representations with mixture of expert neural processes},
  author={Wang, Qi and van Hoof, Herke},
  journal={Advances in neural information processing systems},
  volume={35},
  pages={26242--26255},
  year={2022}
}

@inproceedings{mohseni2023adaptive,
  title={Adaptive conditional quantile neural processes},
  author={Mohseni, Peiman and Duffield, Nick and Mallick, Bani and Hasanzadeh, Arman},
  booktitle={Uncertainty in Artificial Intelligence},
  pages={1445--1455},
  year={2023},
  organization={PMLR}
}

@article{mathieu2023geometric,
  title={Geometric Neural Diffusion Processes},
  author={Mathieu, Emile and Dutordoir, Vincent and Hutchinson, Michael J and De Bortoli, Valentin and Teh, Yee Whye and Turner, Richard E},
  journal={arXiv preprint arXiv:2307.05431},
  year={2023}
}

@inproceedings{wang2020doubly,
  title={Doubly stochastic variational inference for neural processes with hierarchical latent variables},
  author={Wang, Qi and Van Hoof, Herke},
  booktitle={International Conference on Machine Learning},
  pages={10018--10028},
  year={2020},
  organization={PMLR}
}

@inproceedings{wang2022bridge,
  title={Bridge the Inference Gaps of Neural Processes via Expectation Maximization},
  author={Wang, Qi and Federici, Marco and van Hoof, Herke},
  booktitle={The Eleventh International Conference on Learning Representations},
  year={2022}
}

@article{louizos2019functional,
  title={The functional neural process},
  author={Louizos, Christos and Shi, Xiahan and Schutte, Klamer and Welling, Max},
  journal={Advances in Neural Information Processing Systems},
  volume={32},
  year={2019}
}

@article{gupta2022towards,
  title={Towards multi-spatiotemporal-scale generalized pde modeling},
  author={Gupta, Jayesh K and Brandstetter, Johannes},
  journal={arXiv preprint arXiv:2209.15616},
  year={2022}
}

@article{helwig2023group,
  title={Group Equivariant Fourier Neural Operators for Partial Differential Equations},
  author={Helwig, Jacob and Zhang, Xuan and Fu, Cong and Kurtin, Jerry and Wojtowytsch, Stephan and Ji, Shuiwang},
  journal={arXiv preprint arXiv:2306.05697},
  year={2023}
}

@article{gupta2021multiwavelet,
  title={Multiwavelet-based operator learning for differential equations},
  author={Gupta, Gaurav and Xiao, Xiongye and Bogdan, Paul},
  journal={Advances in neural information processing systems},
  volume={34},
  pages={24048--24062},
  year={2021}
}

@article{tran2021factorized,
  title={Factorized fourier neural operators},
  author={Tran, Alasdair and Mathews, Alexander and Xie, Lexing and Ong, Cheng Soon},
  journal={arXiv preprint arXiv:2111.13802},
  year={2021}
}

@article{jha2022neural,
  title={The neural process family: Survey, applications and perspectives},
  author={Jha, Saurav and Gong, Dong and Wang, Xuesong and Turner, Richard E and Yao, Lina},
  journal={arXiv preprint arXiv:2209.00517},
  year={2022}
}

@article{dupont2021generative,
  title={Generative models as distributions of functions},
  author={Dupont, Emilien and Teh, Yee Whye and Doucet, Arnaud},
  journal={arXiv preprint arXiv:2102.04776},
  year={2021}
}

@article{rahman2022u,
  title={U-no: U-shaped neural operators},
  author={Rahman, Md Ashiqur and Ross, Zachary E and Azizzadenesheli, Kamyar},
  journal={arXiv preprint arXiv:2204.11127},
  year={2022}
}

@article{hendrycks2016gaussian,
  title={Gaussian error linear units (gelus)},
  author={Hendrycks, Dan and Gimpel, Kevin},
  journal={arXiv preprint arXiv:1606.08415},
  year={2016}
}

@inproceedings{ronneberger2015u,
  title={U-net: Convolutional networks for biomedical image segmentation},
  author={Ronneberger, Olaf and Fischer, Philipp and Brox, Thomas},
  booktitle={Medical image computing and computer-assisted intervention--MICCAI 2015: 18th international conference, Munich, Germany, October 5-9, 2015, proceedings, part III 18},
  pages={234--241},
  year={2015},
  organization={Springer}
}

@article{feng2022latent,
  title={Latent bottlenecked attentive neural processes},
  author={Feng, Leo and Hajimirsadeghi, Hossein and Bengio, Yoshua and Ahmed, Mohamed Osama},
  journal={arXiv preprint arXiv:2211.08458},
  year={2022}
}

@article{vaswani2017attention,
  title={Attention is all you need},
  author={Vaswani, Ashish and Shazeer, Noam and Parmar, Niki and Uszkoreit, Jakob and Jones, Llion and Gomez, Aidan N and Kaiser, {\L}ukasz and Polosukhin, Illia},
  journal={Advances in neural information processing systems},
  volume={30},
  year={2017}
}

@article{ashman2024translation,
  title={Translation equivariant transformer neural processes},
  author={Ashman, Matthew and Diaconu, Cristiana and Kim, Junhyuck and Sivaraya, Lakee and Markou, Stratis and Requeima, James and Bruinsma, Wessel P and Turner, Richard E},
  journal={arXiv preprint arXiv:2406.12409},
  year={2024}
}

@article{huang2023practical,
  title={Practical equivariances via relational conditional neural processes},
  author={Huang, Daolang and Haussmann, Manuel and Remes, Ulpu and John, ST and Clart{\'e}, Gr{\'e}goire and Luck, Kevin and Kaski, Samuel and Acerbi, Luigi},
  journal={Advances in Neural Information Processing Systems},
  volume={36},
  pages={29201--29238},
  year={2023}
}

@article{kawano2021group,
  title={Group equivariant conditional neural processes},
  author={Kawano, Makoto and Kumagai, Wataru and Sannai, Akiyoshi and Iwasawa, Yusuke and Matsuo, Yutaka},
  journal={arXiv preprint arXiv:2102.08759},
  year={2021}
}

@article{ashman2024approximately,
  title={Approximately Equivariant Neural Processes},
  author={Ashman, Matthew and Diaconu, Cristiana and Weller, Adrian and Bruinsma, Wessel and Turner, Richard},
  journal={Advances in Neural Information Processing Systems},
  volume={37},
  pages={97088--97123},
  year={2024}
}

@inproceedings{holderrieth2021equivariant,
  title={Equivariant learning of stochastic fields: Gaussian processes and steerable conditional neural processes},
  author={Holderrieth, Peter and Hutchinson, Michael J and Teh, Yee Whye},
  booktitle={International conference on machine learning},
  pages={4297--4307},
  year={2021},
  organization={PMLR}
}

@article{field1987relations,
  title={Relations between the statistics of natural images and the response properties of cortical cells},
  author={Field, David J},
  journal={Journal of the Optical Society of America A},
  volume={4},
  number={12},
  pages={2379--2394},
  year={1987},
  publisher={Optical Society of America}
}

@article{wainwright1999scale,
  title={Scale mixtures of Gaussians and the statistics of natural images},
  author={Wainwright, Martin J and Simoncelli, Eero},
  journal={Advances in neural information processing systems},
  volume={12},
  year={1999}
}

@article{ruderman1993statistics,
  title={Statistics of natural images: Scaling in the woods},
  author={Ruderman, Daniel and Bialek, William},
  journal={Advances in neural information processing systems},
  volume={6},
  year={1993}
}

@inproceedings{volpp2021bayesian,
  title={Bayesian Context Aggregation for Neural Processes.},
  author={Volpp, Michael and Fl{\"u}renbrock, Fabian and Grossberger, Lukas and Daniel, Christian and Neumann, Gerhard},
  booktitle={ICLR},
  year={2021}
}

@article{raonic2023convolutional,
  title={Convolutional neural operators for robust and accurate learning of PDEs},
  author={Raonic, Bogdan and Molinaro, Roberto and De Ryck, Tim and Rohner, Tobias and Bartolucci, Francesca and Alaifari, Rima and Mishra, Siddhartha and de B{\'e}zenac, Emmanuel},
  journal={Advances in Neural Information Processing Systems},
  volume={36},
  pages={77187--77200},
  year={2023}
}

@article{bracewell1966fourier,
  title={The Fourier transform and its applications},
  author={Bracewell, Ron and Kahn, Peter B},
  journal={American Journal of Physics},
  volume={34},
  number={8},
  pages={712--712},
  year={1966},
  publisher={American Association of Physics Teachers}
}

@book{oppenheim1999discrete,
  title={Discrete-time signal processing},
  author={Oppenheim, Alan V},
  year={1999},
  publisher={Pearson Education India}
}

@article{loshchilov2017decoupled,
  title={Decoupled weight decay regularization},
  author={Loshchilov, Ilya and Hutter, Frank},
  journal={arXiv preprint arXiv:1711.05101},
  year={2017}
}

@article{Lotka1910contribution,
    author = {Alfred J. Lotka},
    doi = {10.1021/j150111a004},
    issn = {0092-7325},
    journal = {The Journal of Physical Chemistry},
    number = {3},
    pages = {271--274},
    publisher = {American Chemical Society},
    title = {Contribution to the Theory of Periodic Reactions},
    url = {https://doi.org/10.1021/j150111a004},
    volume = {14},
    year = {1910},
}

@article{Volterra1926variazioni,
    author = {V. Volterra},
    title = {Variazioni e Fluttuazioni del Bumero d'Ondividui in Specie Animali Conviventi},
    journal = {Memoria della Reale Accademia Nazionale dei Lincei},
    volume = {2},
    year = {1926},
    pages = {31--113},
}

@article{frigo2005design,
  title={The design and implementation of FFTW3},
  author={Frigo, Matteo and Johnson, Steven G},
  journal={Proceedings of the IEEE},
  volume={93},
  number={2},
  pages={216--231},
  year={2005},
  publisher={IEEE}
}

@article{xiao2024amortized,
  title={Amortized fourier neural operators},
  author={Xiao, Zipeng and Kou, Siqi and Zhongkai, Hao and Lin, Bokai and Deng, Zhijie},
  journal={Advances in Neural Information Processing Systems},
  volume={37},
  pages={115001--115020},
  year={2024}
}

@article{koshizuka2024understanding,
  title={Understanding the expressivity and trainability of fourier neural operator: A mean-field perspective},
  author={Koshizuka, Takeshi and Fujisawa, Masahiro and Tanaka, Yusuke and Sato, Issei},
  journal={Advances in Neural Information Processing Systems},
  volume={37},
  pages={11021--11060},
  year={2024}
}

@article{bartolucci2023representation,
  title={Representation equivalent neural operators: a framework for alias-free operator learning},
  author={Bartolucci, Francesca and de Bezenac, Emmanuel and Raonic, Bogdan and Molinaro, Roberto and Mishra, Siddhartha and Alaifari, Rima},
  journal={Advances in Neural Information Processing Systems},
  volume={36},
  pages={69661--69672},
  year={2023}
}

@article{xiao2023improved,
  title={Improved operator learning by orthogonal attention},
  author={Xiao, Zipeng and Hao, Zhongkai and Lin, Bokai and Deng, Zhijie and Su, Hang},
  journal={arXiv preprint arXiv:2310.12487},
  year={2023}
}

@article{zheng2024alias,
  title={Alias-Free Mamba Neural Operator},
  author={Zheng, Jianwei and Li, Wei and Xu, Ni and Zhu, Junwei and Zhang, Xiaoqin},
  journal={Advances in Neural Information Processing Systems},
  volume={37},
  pages={52962--52995},
  year={2024}
}

@article{luo2024hierarchical,
  title={Hierarchical neural operator transformer with learnable frequency-aware loss prior for arbitrary-scale super-resolution},
  author={Luo, Xihaier and Qian, Xiaoning and Yoon, Byung-Jun},
  journal={arXiv preprint arXiv:2405.12202},
  year={2024}
}

@article{qin2024toward,
  title={Toward a Better Understanding of Fourier Neural Operators from a Spectral Perspective},
  author={Qin, Shaoxiang and Lyu, Fuyuan and Peng, Wenhui and Geng, Dingyang and Wang, Ju and Tang, Xing and Leroyer, Sylvie and Gao, Naiping and Liu, Xue and Wang, Liangzhu Leon},
  journal={arXiv preprint arXiv:2404.07200},
  year={2024}
}

@inproceedings{liu2023largest,
  title={LargeST: A Benchmark Dataset for Large-Scale Traffic Forecasting},
  author={Liu, Xu and Xia, Yutong and Liang, Yuxuan and Hu, Junfeng and Wang, Yiwei and Bai, Lei and Huang, Chao and Liu, Zhenguang and Hooi, Bryan and Zimmermann, Roger},
  booktitle={Advances in Neural Information Processing Systems},
  year={2023}
}

@inproceedings{
ashman2025gridded,
title={Gridded Transformer Neural Processes for Spatio-Temporal Data},
author={Matthew Ashman and Cristiana Diaconu and Eric Langezaal and Adrian Weller and Richard E. Turner},
booktitle={Forty-second International Conference on Machine Learning},
year={2025},
url={https://openreview.net/forum?id=O0oe7hPtbl}
}

@article{allen2025end,
  title={End-to-end data-driven weather prediction},
  author={Allen, Anna and Markou, Stratis and Tebbutt, Will and Requeima, James and Bruinsma, Wessel P and Andersson, Tom R and Herzog, Michael and Lane, Nicholas D and Chantry, Matthew and Hosking, J Scott and others},
  journal={Nature},
  volume={641},
  number={8065},
  pages={1172--1179},
  year={2025},
  publisher={Nature Publishing Group}
}

@article{bruinsma2024convolutional,
  title={Convolutional Conditional Neural Processes},
  author={Bruinsma, Wessel P},
  journal={arXiv preprint arXiv:2408.09583},
  year={2024}
}

@article{vaughan2021convolutional,
  title={Convolutional conditional neural processes for local climate downscaling},
  author={Vaughan, Anna and Tebbutt, Will and Hosking, J Scott and Turner, Richard E},
  journal={Geoscientific Model Development Discussions},
  volume={2021},
  pages={1--25},
  year={2021},
  publisher={G{\"o}ttingen, Germany}
}

@article{xu2023deep,
  title={Deep stochastic processes via functional markov transition operators},
  author={Xu, Jin and Dupont, Emilien and M{\"a}rtens, Kaspar and Rainforth, Thomas and Teh, Yee Whye},
  journal={Advances in Neural Information Processing Systems},
  volume={36},
  pages={37975--37994},
  year={2023}
}

@inproceedings{he2016deep,
  title={Deep residual learning for image recognition},
  author={He, Kaiming and Zhang, Xiangyu and Ren, Shaoqing and Sun, Jian},
  booktitle={Proceedings of the IEEE conference on computer vision and pattern recognition},
  pages={770--778},
  year={2016}
}

@article{fukushima1980neocognitron,
  title={Neocognitron: A self-organizing neural network model for a mechanism of pattern recognition unaffected by shift in position},
  author={Fukushima, Kunihiko},
  journal={Biological cybernetics},
  volume={36},
  number={4},
  pages={193--202},
  year={1980},
  publisher={Springer}
}

@article{lecun1989backpropagation,
  title={Backpropagation applied to handwritten zip code recognition},
  author={LeCun, Yann and Boser, Bernhard and Denker, John S and Henderson, Donnie and Howard, Richard E and Hubbard, Wayne and Jackel, Lawrence D},
  journal={Neural computation},
  volume={1},
  number={4},
  pages={541--551},
  year={1989},
  publisher={MIT Press}
}

@article{chen1995universal,
  title={Universal approximation to nonlinear operators by neural networks with arbitrary activation functions and its application to dynamical systems},
  author={Chen, Tianping and Chen, Hong},
  journal={IEEE transactions on neural networks},
  volume={6},
  number={4},
  pages={911--917},
  year={1995},
  publisher={IEEE}
}

@article{kovachki2021universal,
  title={On universal approximation and error bounds for Fourier neural operators},
  author={Kovachki, Nikola and Lanthaler, Samuel and Mishra, Siddhartha},
  journal={Journal of Machine Learning Research},
  volume={22},
  number={290},
  pages={1--76},
  year={2021}
}

@article{zaheer2017deep,
  title={Deep sets},
  author={Zaheer, Manzil and Kottur, Satwik and Ravanbakhsh, Siamak and Poczos, Barnabas and Salakhutdinov, Russ R and Smola, Alexander J},
  journal={Advances in neural information processing systems},
  volume={30},
  year={2017}
}

@inproceedings{xu2020metafun,
  title={Metafun: Meta-learning with iterative functional updates},
  author={Xu, Jin and Ton, Jean-Francois and Kim, Hyunjik and Kosiorek, Adam and Teh, Yee Whye},
  booktitle={International Conference on Machine Learning},
  pages={10617--10627},
  year={2020},
  organization={PMLR}
}

@inproceedings{wagstaff2019limitations,
  title={On the limitations of representing functions on sets},
  author={Wagstaff, Edward and Fuchs, Fabian and Engelcke, Martin and Posner, Ingmar and Osborne, Michael A},
  booktitle={International conference on machine learning},
  pages={6487--6494},
  year={2019},
  organization={PMLR}
}

@article{bloem2020probabilistic,
  title={Probabilistic symmetries and invariant neural networks},
  author={Bloem-Reddy, Benjamin and Teh, Yee Whye},
  journal={Journal of Machine Learning Research},
  volume={21},
  number={90},
  pages={1--61},
  year={2020}
}

@inproceedings{qi2017pointnet,
  title={Pointnet: Deep learning on point sets for 3d classification and segmentation},
  author={Qi, Charles R and Su, Hao and Mo, Kaichun and Guibas, Leonidas J},
  booktitle={Proceedings of the IEEE conference on computer vision and pattern recognition},
  pages={652--660},
  year={2017}
}

@inproceedings{
knigge2023modelling,
title={Modelling Long Range Dependencies in \$N\$D: From Task-Specific to a General Purpose {CNN}},
author={David M Knigge and David W. Romero and Albert Gu and Efstratios Gavves and Erik J Bekkers and Jakub Mikolaj Tomczak and Mark Hoogendoorn and Jan-jakob Sonke},
booktitle={The Eleventh International Conference on Learning Representations },
year={2023},
url={https://openreview.net/forum?id=ZW5aK4yCRqU}
}

@inproceedings{ding2022scaling,
  title={Scaling up your kernels to 31x31: Revisiting large kernel design in cnns},
  author={Ding, Xiaohan and Zhang, Xiangyu and Han, Jungong and Ding, Guiguang},
  booktitle={Proceedings of the IEEE/CVF conference on computer vision and pattern recognition},
  pages={11963--11975},
  year={2022}
}

@InProceedings{Peng_2017_CVPR,
author = {Peng, Chao and Zhang, Xiangyu and Yu, Gang and Luo, Guiming and Sun, Jian},
title = {Large Kernel Matters -- Improve Semantic Segmentation by Global Convolutional Network},
booktitle = {Proceedings of the IEEE Conference on Computer Vision and Pattern Recognition (CVPR)},
month = {July},
year = {2017}
}

@article{luo2016understanding,
  title={Understanding the effective receptive field in deep convolutional neural networks},
  author={Luo, Wenjie and Li, Yujia and Urtasun, Raquel and Zemel, Richard},
  journal={Advances in neural information processing systems},
  volume={29},
  year={2016}
}

@inproceedings{wang2018non,
  title={Non-local neural networks},
  author={Wang, Xiaolong and Girshick, Ross and Gupta, Abhinav and He, Kaiming},
  booktitle={Proceedings of the IEEE conference on computer vision and pattern recognition},
  pages={7794--7803},
  year={2018}
}

@article{ramachandran2019stand,
  title={Stand-alone self-attention in vision models},
  author={Ramachandran, Prajit and Parmar, Niki and Vaswani, Ashish and Bello, Irwan and Levskaya, Anselm and Shlens, Jon},
  journal={Advances in neural information processing systems},
  volume={32},
  year={2019}
}

@inproceedings{wang2020axial,
  title={Axial-deeplab: Stand-alone axial-attention for panoptic segmentation},
  author={Wang, Huiyu and Zhu, Yukun and Green, Bradley and Adam, Hartwig and Yuille, Alan and Chen, Liang-Chieh},
  booktitle={European conference on computer vision},
  pages={108--126},
  year={2020},
  organization={Springer}
}

@article{shaw2018self,
  title={Self-attention with relative position representations},
  author={Shaw, Peter and Uszkoreit, Jakob and Vaswani, Ashish},
  journal={arXiv preprint arXiv:1803.02155},
  year={2018}
}

@article{su2024roformer,
  title={Roformer: Enhanced transformer with rotary position embedding},
  author={Su, Jianlin and Ahmed, Murtadha and Lu, Yu and Pan, Shengfeng and Bo, Wen and Liu, Yunfeng},
  journal={Neurocomputing},
  volume={568},
  pages={127063},
  year={2024},
  publisher={Elsevier}
}

@article{liu2023domain,
  title={Domain agnostic fourier neural operators},
  author={Liu, Ning and Jafarzadeh, Siavash and Yu, Yue},
  journal={Advances in Neural Information Processing Systems},
  volume={36},
  pages={47438--47450},
  year={2023}
}

@article{tripura2023wavelet,
  title={Wavelet Neural Operator for solving parametric partial differential equations in computational mechanics problems},
  author={Tripura, Tapas and Chakraborty, Souvik},
  journal={Computer Methods in Applied Mechanics and Engineering},
  volume={404},
  pages={115783},
  year={2023},
  publisher={Elsevier}
}

@inproceedings{li2024multi,
  title={Multi-resolution active learning of Fourier neural operators},
  author={Li, Shibo and Yu, Xin and Xing, Wei and Kirby, Robert and Narayan, Akil and Zhe, Shandian},
  booktitle={International Conference on Artificial Intelligence and Statistics},
  pages={2440--2448},
  year={2024},
  organization={PMLR}
}

@article{zaheer2020big,
  title={Big bird: Transformers for longer sequences},
  author={Zaheer, Manzil and Guruganesh, Guru and Dubey, Kumar Avinava and Ainslie, Joshua and Alberti, Chris and Ontanon, Santiago and Pham, Philip and Ravula, Anirudh and Wang, Qifan and Yang, Li and others},
  journal={Advances in neural information processing systems},
  volume={33},
  pages={17283--17297},
  year={2020}
}

@article{beltagy2020longformer,
  title={Longformer: The long-document transformer},
  author={Beltagy, Iz and Peters, Matthew E and Cohan, Arman},
  journal={arXiv preprint arXiv:2004.05150},
  year={2020}
}

@inproceedings{liu2021swin,
  title={Swin transformer: Hierarchical vision transformer using shifted windows},
  author={Liu, Ze and Lin, Yutong and Cao, Yue and Hu, Han and Wei, Yixuan and Zhang, Zheng and Lin, Stephen and Guo, Baining},
  booktitle={Proceedings of the IEEE/CVF international conference on computer vision},
  pages={10012--10022},
  year={2021}
}

@article{ding2023longnet,
  title={Longnet: Scaling transformers to 1,000,000,000 tokens},
  author={Ding, Jiayu and Ma, Shuming and Dong, Li and Zhang, Xingxing and Huang, Shaohan and Wang, Wenhui and Zheng, Nanning and Wei, Furu},
  journal={arXiv preprint arXiv:2307.02486},
  year={2023}
}

@inproceedings{rahaman2019spectral,
  title={On the spectral bias of neural networks},
  author={Rahaman, Nasim and Baratin, Aristide and Arpit, Devansh and Draxler, Felix and Lin, Min and Hamprecht, Fred and Bengio, Yoshua and Courville, Aaron},
  booktitle={International conference on machine learning},
  pages={5301--5310},
  year={2019},
  organization={PMLR}
}

@inproceedings{
vasudeva2025transformers,
title={Transformers Learn Low Sensitivity Functions: Investigations and Implications},
author={Bhavya Vasudeva and Deqing Fu and Tianyi Zhou and Elliott Kau and Youqi Huang and Vatsal Sharan},
booktitle={The Thirteenth International Conference on Learning Representations},
year={2025},
url={https://openreview.net/forum?id=4ikjWBs3tE}
}

@article{ronen2019convergence,
  title={The convergence rate of neural networks for learned functions of different frequencies},
  author={Ronen, Basri and Jacobs, David and Kasten, Yoni and Kritchman, Shira},
  journal={Advances in Neural Information Processing Systems},
  volume={32},
  year={2019}
}

@article{tancik2020fourier,
  title={Fourier features let networks learn high frequency functions in low dimensional domains},
  author={Tancik, Matthew and Srinivasan, Pratul and Mildenhall, Ben and Fridovich-Keil, Sara and Raghavan, Nithin and Singhal, Utkarsh and Ramamoorthi, Ravi and Barron, Jonathan and Ng, Ren},
  journal={Advances in neural information processing systems},
  volume={33},
  pages={7537--7547},
  year={2020}
}

@article{fridovich2022spectral,
  title={Spectral bias in practice: The role of function frequency in generalization},
  author={Fridovich-Keil, Sara and Gontijo Lopes, Raphael and Roelofs, Rebecca},
  journal={Advances in Neural Information Processing Systems},
  volume={35},
  pages={7368--7382},
  year={2022}
}

@inproceedings{basri2020frequency,
  title={Frequency bias in neural networks for input of non-uniform density},
  author={Basri, Ronen and Galun, Meirav and Geifman, Amnon and Jacobs, David and Kasten, Yoni and Kritchman, Shira},
  booktitle={International conference on machine learning},
  pages={685--694},
  year={2020},
  organization={PMLR}
}

@inproceedings{cimpoi2014describing,
  title={Describing textures in the wild},
  author={Cimpoi, Mircea and Maji, Subhransu and Kokkinos, Iasonas and Mohamed, Sammy and Vedaldi, Andrea},
  booktitle={Proceedings of the IEEE conference on computer vision and pattern recognition},
  pages={3606--3613},
  year={2014}
}

@article{mackay1992practical,
  title={A practical Bayesian framework for backpropagation networks},
  author={MacKay, David JC},
  journal={Neural computation},
  volume={4},
  number={3},
  pages={448--472},
  year={1992},
  publisher={MIT Press One Rogers Street, Cambridge, MA 02142-1209, USA journals-info~…}
}

@article{wolpert1993bayesian,
  title={Bayesian backpropagation over io functions rather than weights},
  author={Wolpert, David H},
  journal={Advances in neural information processing systems},
  volume={6},
  year={1993}
}

@article{qiu2023should,
  title={Should we learn most likely functions or parameters?},
  author={Qiu, Shikai and Rudner, Tim GJ and Kapoor, Sanyam and Wilson, Andrew G},
  journal={Advances in Neural Information Processing Systems},
  volume={36},
  pages={35814--35835},
  year={2023}
}

@inproceedings{rudner2023function,
  title={Function-space regularization in neural networks: A probabilistic perspective},
  author={Rudner, Tim GJ and Kapoor, Sanyam and Qiu, Shikai and Wilson, Andrew Gordon},
  booktitle={International Conference on Machine Learning},
  pages={29275--29290},
  year={2023},
  organization={PMLR}
}

@inproceedings{
wu2024functional,
title={Functional Wasserstein Bridge Inference for Bayesian Deep Learning},
author={Mengjing Wu and Junyu Xuan and Jie Lu},
booktitle={The 40th Conference on Uncertainty in Artificial Intelligence},
year={2024},
url={https://openreview.net/forum?id=Wnht2IqzlN}
}

@article{phillips2022spectral,
  title={Spectral diffusion processes},
  author={Phillips, Angus and Seror, Thomas and Hutchinson, Michael and De Bortoli, Valentin and Doucet, Arnaud and Mathieu, Emile},
  journal={arXiv preprint arXiv:2209.14125},
  year={2022}
}

@article{franzese2023continuous,
  title={Continuous-time functional diffusion processes},
  author={Franzese, Giulio and Corallo, Giulio and Rossi, Simone and Heinonen, Markus and Filippone, Maurizio and Michiardi, Pietro},
  journal={Advances in Neural Information Processing Systems},
  volume={36},
  pages={37370--37400},
  year={2023}
}

@article{calvello2024continuum,
  title={Continuum attention for neural operators},
  author={Calvello, Edoardo and Kovachki, Nikola B and Levine, Matthew E and Stuart, Andrew M},
  journal={arXiv preprint arXiv:2406.06486},
  year={2024}
}

@article{dugas2000incorporating,
  title={Incorporating second-order functional knowledge for better option pricing},
  author={Dugas, Charles and Bengio, Yoshua and B{\'e}lisle, Fran{\c{c}}ois and Nadeau, Claude and Garcia, Ren{\'e}},
  journal={Advances in neural information processing systems},
  volume={13},
  year={2000}
}

@inproceedings{nair2010rectified,
  title={Rectified linear units improve restricted boltzmann machines},
  author={Nair, Vinod and Hinton, Geoffrey E},
  booktitle={Proceedings of the 27th international conference on machine learning (ICML-10)},
  pages={807--814},
  year={2010}
}

@article{gardner2018gpytorch,
  title={Gpytorch: Blackbox matrix-matrix gaussian process inference with gpu acceleration},
  author={Gardner, Jacob and Pleiss, Geoff and Weinberger, Kilian Q and Bindel, David and Wilson, Andrew G},
  journal={Advances in neural information processing systems},
  volume={31},
  year={2018}
}

@article{paszke2019pytorch,
  title={Pytorch: An imperative style, high-performance deep learning library},
  author={Paszke, Adam and Gross, Sam and Massa, Francisco and Lerer, Adam and Bradbury, James and Chanan, Gregory and Killeen, Trevor and Lin, Zeming and Gimelshein, Natalia and Antiga, Luca and others},
  journal={Advances in neural information processing systems},
  volume={32},
  year={2019}
}

@inproceedings{pascanu2013difficulty,
  title={On the difficulty of training recurrent neural networks},
  author={Pascanu, Razvan and Mikolov, Tomas and Bengio, Yoshua},
  booktitle={International conference on machine learning},
  pages={1310--1318},
  year={2013},
  organization={Pmlr}
}

@InProceedings{pmlr-v238-kerrigan24a,
  title = 	 {Functional Flow Matching},
  author =       {Kerrigan, Gavin and Migliorini, Giosue and Smyth, Padhraic},
  booktitle = 	 {Proceedings of The 27th International Conference on Artificial Intelligence and Statistics},
  pages = 	 {3934--3942},
  year = 	 {2024},
  editor = 	 {Dasgupta, Sanjoy and Mandt, Stephan and Li, Yingzhen},
  volume = 	 {238},
  series = 	 {Proceedings of Machine Learning Research},
  month = 	 {02--04 May},
  publisher =    {PMLR},
  url = 	 {https://proceedings.mlr.press/v238/kerrigan24a.html},
}

@article{zhang2025flow,
  title={Flow Straight and Fast in Hilbert Space: Functional Rectified Flow},
  author={Zhang, Jianxin and Scott, Clayton},
  journal={arXiv preprint arXiv:2509.10384},
  year={2025}
}

@inproceedings{
lim2023score,
title={Score-based Generative Modeling through Stochastic Evolution Equations in Hilbert Spaces},
author={Sungbin Lim and Eunbi Yoon and Taehyun Byun and Taewon Kang and Seungwoo Kim and Kyungjae Lee and Sungjoon Choi},
booktitle={Thirty-seventh Conference on Neural Information Processing Systems},
year={2023},
url={https://openreview.net/forum?id=GrElRvXnEj}
}

@article{lim2022energy,
  title={Energy-based models for functional data using path measure tilting},
  author={Lim, Jen Ning and Vollmer, Sebastian and Wolf, Lorenz and Duncan, Andrew},
  journal={arXiv preprint arXiv:2202.01929},
  year={2022}
}

@inproceedings{yang2020energy,
  title={Energy-based processes for exchangeable data},
  author={Yang, Mengjiao and Dai, Bo and Dai, Hanjun and Schuurmans, Dale},
  booktitle={International Conference on Machine Learning},
  pages={10681--10692},
  year={2020},
  organization={PMLR}
}
\bibliographystyle{plainnat}

\newpage
\appendix

\section{Discretization Sensitivity of the DFT}\label{appensix: DFT-discretization}
To illustrate the sensitivity of the discrete Fourier transform (DFT) to the underlying discretization—and how changes in grid geometry can degrade predictive performance—we consider a simple one-dimensional example. Let $0 < \Delta \ll 1$ denote a fixed discretization resolution, interpreted as the spacing between adjacent grid points. Define the grid
\[
\mathcal{G}_1 = (x_0, \dots, x_{m_1-1})
\]
as a uniform discretization of the interval $[0,1]$, where $m_1 = \lfloor 1/\Delta \rfloor$ and the points are ordered such that $x_{n} < x_{n+1}$ for all $n \in \{0,\dots,m_1-2\}$. Let $h := \varphi_e[\mathcal{D}_c]$ denote the encoded latent function, and let its sampled values on $\mathcal{G}_1$ be
\[
\bigl(h(x)\bigr)_{x \in \mathcal{G}_1}
= \bigl(h(x_0), \dots, h(x_{m_1-1})\bigr).
\]
The DFT of $h$ on this grid is given by
\[
\hat{\mathcal{F}}\!\left\{ \bigl(h(x)\bigr)_{x \in \mathcal{G}_1} \right\}(k)
=
\sum_{n=0}^{m_1-1}
h(x_n)\, e^{-i 2\pi \frac{k}{m_1} n},
\qquad
k \in \{0,\dots,m_1-1\}.
\]
This yields Fourier coefficients at the normalized frequencies
\[
\Xi_1 = \left( \tfrac{k}{m_1} \right)_{k=0}^{m_1-1}.
\]
As discussed in Section~\ref{subsec: FNOs}, the Fourier Neural Operator (FNO) implicitly ties its kernel parameterization to the specific frequency set $\Xi_1$ encountered during training. Now suppose the trained model is evaluated on a larger spatial domain, e.g.\ $[0,2]$, while maintaining the same resolution $\Delta$. The corresponding grid
\[
\mathcal{G}_2 = (x'_0, \dots, x'_{m_2-1})
\]
has size $m_2 = \lfloor 2/\Delta \rfloor$, and the associated DFT produces coefficients at normalized frequencies
\[
\Xi_2 = \left( \tfrac{k}{m_2} \right)_{k=0}^{m_2-1}.
\]
Since $m_2 \ge 2m_1$, we have $|\Xi_2| \ge 2|\Xi_1|$, and, crucially, the frequency locations themselves differ. Let $1 \le k_{\max} \le m_1$ denote the number of Fourier modes retained during training. Even when restricting attention to the lowest $k_{\max}$ modes, the frequencies $\{k/m_2\}_{k=0}^{k_{\max}-1}$ do not coincide with $\{k/m_1\}_{k=0}^{k_{\max}-1}$. This misalignment induces a spectral mismatch between training and evaluation, illustrating how changes in discretization—even at fixed resolution—can disrupt the learned Fourier parameterization and lead to degraded generalization.

\section{Experimental Details}\label{appendix: experimental-details}
All implementations are written in PyTorch \citep{paszke2019pytorch} and publicly available at \href{https://github.com/peiman-m/SConvCNP}{https://github.com/peiman-m/SConvCNP}. We used a single NVIDIA A100 GPU with 40 GB of memory for all the computations. Our code is based on the implementations of \citet{ashman2024translation} and \citet{bruinsma2023autoregressive}.

\subsection{Synthetic 1-D Regression}\label{appendix: synthetic-regression-experiment-details}

\subsubsection{Model Architectures}
\label{appendix: synthetic-regression-model-architectures}

This section details the architectures of all CNPs variants used in our experiments. Each model outputs a Gaussian predictive distribution parameterized by a mean and a pre-softplus \citep{dugas2000incorporating} scale. The pre-softplus value is passed through a softplus transformation, and a minimum noise term of $10^{-6}$ is then added to the resulting scale. The Gaussian factorizes across query points---as is standard for CNPs (see Section~\ref{subsec: NPs})---and, for multidimensional outputs, across output dimensions as well. Thus, predictions assume independence both across query locations and across components of each output vector. Unless otherwise noted, all nonlinearities use ReLU activations \citep{nair2010rectified}.

\paragraph{Conditional Neural Process (CNP):}
The CNP encodes each context pair $(x_{c,k}, y_{c,k}) \in \mathcal{D}_c$ using separate input and output pathways. The input $x_{c,k}$ and output $y_{c,k}$ are first passed through distinct MLPs with two hidden layers each of dimension 256, producing 256-dimensional representations $\varepsilon^{(x)}_{c,k}$ and $\varepsilon^{(y)}_{c,k}$. Their concatenation $\varepsilon_{c,k} = [\varepsilon^{(x)}_{c,k}, \varepsilon^{(y)}_{c,k}]$ is then processed by a deeper MLP with six hidden layers of width 256 and a final 256-dimensional output layer, yielding an embedding for each context pair. Context embeddings $\varepsilon_{c, k}$ are averaged to form a single aggregated representation $\varepsilon_{c}$ of dimension 256. For prediction, each query input $x_{q,l}$ is concatenated with the context representation $\varepsilon_{c}$ and passed through a decoder MLP consisting of six hidden layers of dimension 256. The decoder’s final output layer has dimensionality $2 d_y$, parameterizing the mean and the log-scale parameters of the Gaussian predictive distribution.

\paragraph{Attentive Conditional Neural Process (AttCNP).}
The AttCNP implementation follows the deterministic architecture introduced by \citet{kim2019attentive}. The initial encoding stage mirrors that of the CNP: each context pair $(x_{c,k}, y_{c,k}) \in \mathcal{D}_c$ is processed by two separate MLPs (each with two hidden layers of width 256), producing embeddings $\varepsilon^{(x)}_{c,k}$ and $\varepsilon^{(y)}_{c,k}$. These are concatenated to form $\varepsilon_{c,k} = [\varepsilon^{(x)}_{c,k}, \varepsilon^{(y)}_{c,k}]$, which is subsequently passed through an additional two-layer MLP with hidden width 256. Departing from the CNP, the AttCNP applies self-attention to the set of context embeddings. We use two layers of multi-head self-attention, each with 8 heads (head dimension 32), a feedforward subnetwork with hidden width 256, residual connections, and pre-layer normalization for both the attention and feedforward blocks. The input and output dimensionality of each attention layer is fixed at 256. Query representations $\varepsilon_{q,l}$ are computed using a single layer of multi-head cross-attention between the (attended) context embeddings and the query embeddings, with the same configuration as above (8 heads of dimension~32, feedforward width~256, residual connections, and pre-normalization). The decoder is an MLP with six hidden layers of width~256. Its final layer outputs $2 d_y$ units, corresponding to the mean and log-scale parameters of a Gaussian predictive distribution.

\paragraph{Transformer Neural Process (TNP).}
To apply the TNP, we first construct token representations for both context and query points. 
For each context pair $(x_{c,k}, y_{c,k}) \in \mathcal{D}_c$, we form the token
\[
[x_{c,k},\, y_{c,k},\, 1],
\]
where the final singleton ``1'' serves as a density flag indicating that the observation $y_{c,k}$ at $x_{c,k}$ is available. 
For each query location $x_{q,l}$, we instead construct
\[
[x_{q,l},\, \mathbf{0},\, 0], 
\]
where the dummy zero-vector matches the shape of $y_{c,k}$ and the final ``0'' indicates the absence of an observation~\citep{nguyen2022transformer, ashman2024translation}. 
All tokens are then passed through a shared two-layer MLP (width 256), producing initial embeddings $\varepsilon_{c,k}$ and $\varepsilon_{q,l}$.

In the original TNP architecture~\citep{nguyen2022transformer}, context and query embeddings are processed jointly by a transformer encoder. An attention mask prevents (i) interactions among queries and (ii) context$\rightarrow$query attention, ensuring that context representations remain query-independent, while queries may attend to contexts. This design incurs quadratic complexity in the total sequence length, 
\[
    \mathcal{O}((|\mathcal{D}_c| + |\mathcal{D}_q|)^2).
\]
Because queries never attend to one another, \citet{feng2022latent} introduced the Efficient Query TNP, which we adopt here.
This variant first applies self-attention to context tokens to produce updated context embeddings, then processes each query using cross-attention over these updated contexts.
The resulting two-branch structure (contexts processed twice, queries once) reduces the overall complexity to
\[
    \mathcal{O}\bigl(|\mathcal{D}_c|^2 + |\mathcal{D}_q|\,|\mathcal{D}_c|\bigr).
\]

Our model uses six transformer layers with eight attention heads (head dimension~32), each followed by a feedforward subnetwork of width~256. We use the standard pre-norm architecture, applying layer normalization before both the attention and feedforward modules. All attention operations share a 256-dimensional embedding dimension. At each layer, we first apply context self-attention and then query--context cross-attention. Notably, the same multi-head attention parameters are used for both operations; the distinction arises only from which embeddings serve as queries, keys, and values (yielding self-attention when they coincide, and cross-attention otherwise). This is not a modeling requirement---separate parameter sets could be used, as demonstrated in Appendix~\ref{appendix: image-completion-model-architectures}.
The final-layer query embeddings are passed to a decoder MLP with two hidden layers (width~256), whose output parameterizes a Gaussian predictive distribution via $2 d_y$ units corresponding to the mean and log-scale.

\paragraph{Translation-Equivariant Transformer Neural Process (TE-TNP).}
The token construction in the TE-TNP closely follows that of the TNP, with a crucial modification to ensure translation equivariance: input locations are excluded from the token representations.
Specifically, for each context pair $(x_{c,k}, y_{c,k}) \in \mathcal{D}_c$, we construct the token
\[
[y_{c,k},\, 1],
\]
while for each query location $x_{q,l}$ we use the token
\[
[\mathbf{0},\, 0].
\]
All tokens are passed through a shared two-layer MLP with width 256 to produce initial embeddings $\varepsilon_{c,k}$ and $\varepsilon_{q,l}$.

The TE-TNP replaces the standard multi-head attention mechanism with the translation-equivariant attention proposed by \citet{ashman2024translation}. For each attention head, we first compute the matrix of pairwise scaled dot products between token embeddings, analogous to conventional attention but independent of absolute input locations. In parallel, we compute the matrix of pairwise \emph{differences} between token locations. We then concatenate the scaled dot-product scores from all heads with the corresponding pairwise location differences. This augmented pairwise similarity representation is processed by an MLP—acting as an implicit kernel—with two hidden layers of width 256 and an output dimension equal to the number of attention heads, yielding the final translation-equivariant attention logits.

Apart from this modified attention module, the remainder of the architecture and computational pipeline follows that of the TNP.

\paragraph{Convolutional Conditional Neural Process (ConvCNP):}
For the ConvCNP, we begin by determining, for each input dimension, the minimum and maximum coordinates observed across both the context and query sets. These extrema are expanded by a small margin of~0.1 in each dimension. The interval between the expanded minima and maxima is then uniformly discretized at a resolution of 64~points per unit. When necessary---for example, to satisfy the CNN’s minimum grid-size requirements---the discretization range is further enlarged while maintaining the same resolution. The resulting one-dimensional grids are combined via a Cartesian product to yield a uniform grid~$\mathcal{G}$.

The functional embedding in \eqref{eq: functional-embedding} is evaluated on this grid. The Gaussian kernels used in this embedding are initialized with per-dimension length scales set to twice the grid resolution, i.e., $2/64$. For multi-dimensional inputs and outputs, the model uses separate length scales for each dimension. Following \citet{bruinsma2023autoregressive}, the embedding is further normalized using a \emph{density channel}, defined as
\[
    \mathrm{Density}(x) 
    = 
    \sum_{(x_c,\, y_c)\in\mathcal{D}_c}
      \psi_e(x - x_c).
\]
The functional and density channels are concatenated to form the final grid representation
\begin{equation}\label{eq: function+density}
    \left(
    \mathrm{Density}(x_g),\ 
    \frac{
        \sum_{(x_c,\, y_c) \in \mathcal{D}_c}
        \phi(y_c)\,\psi_e(x_g - x_c)
    }{
        \mathrm{Density}(x_g)
    }
    \right)_{x_g \in \mathcal{G}}.
\end{equation}

Each grid point is processed independently by an MLP with two hidden layers of width~128. The resulting features are passed to a CNN based on a U-Net architecture \citep{ronneberger2015u}, consisting of six residual convolutional blocks (kernel size~11, stride~2, 128 channels) in the encoder, followed by a symmetric sequence of six transposed convolutional blocks in the decoder, with skip connections following the design of \citet{bruinsma2023autoregressive}. Since the U-Net downsamples the spatial resolution by a factor of~64, we ensure that the constructed grid size is divisible by~64. For the one-dimensional benchmarks considered in this work, this is achieved by symmetrically enlarging the discretization interval when necessary so that the CNN output aligns exactly with the input grid.

To obtain predictive distribution parameters at query locations---which may not lie on the grid $\mathcal{G}$--- an interpolation scheme analogous to \eqref{eq: function+density} is applied, except that the weighted sum over the feature-map values is not normalized. For this interpolation, we employ a separate Gaussian kernel with learnable length-scale parameters, distinct from the kernel used in the functional embedding. The resulting query-specific embeddings are then processed by a decoder MLP with two hidden layers (width~128), whose output consists of $2 d_y$ units parameterizing the mean and log-scale of a Gaussian predictive distribution.

\paragraph{Spectral Convolutional Conditional Neural Process (SConvCNP)}
The SConvCNP replaces the ConvCNP's U-Net backbone with a U-shaped Fourier Neural Operator (FNO) architecture~\citep{li2020fourier, rahman2022u}. A residual Fourier block is denoted
\[
\mathrm{F}(c_{\mathrm{in}},\, c_{\mathrm{out}},\, s_{\mathrm{in}},\, s_{\mathrm{out}},\, m_{\mathrm{f}}),
\]
where $c_{\mathrm{in}}$ and $c_{\mathrm{out}}$ are channel dimensions, $s_{\mathrm{in}}$ and $s_{\mathrm{out}}$ are spatial sizes, and $m_{\mathrm{f}}$ is the number of retained Fourier modes.

The representation from \eqref{eq: function+density}, augmented with positional encodings (grid coordinates), is first processed by an MLP with one hidden layer of width~64 and GELU activations \citep{hendrycks2016gaussian}. Its output is then passed through the following sequence of residual Fourier blocks:
\begin{itemize}
    \item $\mathrm{L_1} = \mathrm{F}(64,\,128,\,|\mathcal{G}|,\,\lfloor |\mathcal{G}|/2 \rfloor,\,32)$
    \item $\mathrm{L_2} = \mathrm{F}(128,\,128,\,\lfloor |\mathcal{G}|/2 \rfloor,\,\lfloor |\mathcal{G}|/4 \rfloor,\,32)$
    \item $\mathrm{L_3} = \mathrm{F}(128,\,256,\,\lfloor |\mathcal{G}|/4 \rfloor,\,32)$
    \item $\mathrm{L_4} = \mathrm{F}(256,\,128,\,\lfloor |\mathcal{G}|/2 \rfloor,\,32)$
    \item $\mathrm{L_5} = \mathrm{F}(256,\,128,\,|\mathcal{G}|,\,32)$
\end{itemize}

Layers $\mathrm{L_1} \rightarrow \mathrm{L_2} \rightarrow \mathrm{L_3} \rightarrow \mathrm{L_4}$ form the contractive–expansive path. The final block $\mathrm{L_5}$ receives the channel-wise concatenation of the outputs of $\mathrm{L_4}$ and $\mathrm{L_1}$, yielding the U-shaped skip connection. Its output is concatenated with the initial MLP features and processed by a final MLP (one hidden layer, width~128, GELU) to produce the SConvCNP representation for the decoder.

\paragraph{Parameter Count.}
Table \ref{table: synthetic-regression-parameter-count} summarizes the number of learnable parameters for all models used in our 1D synthetic regression experiments. 

\begin{table*}[!ht]\centering
    \caption{Learnable parameter counts for all models used in the 1D synthetic regression experiments.}
    \vspace{2.5pt}
    \renewcommand{\arraystretch}{1.3}
    \label{table: synthetic-regression-parameter-count}
    \scalebox{1.0}{
    \begin{tabular}{l@{\hskip 15pt} c@{\hskip 10pt} c@{\hskip 10pt} c@{\hskip 10pt} c@{\hskip 10pt} c@{\hskip 10pt} c@{\hskip 1pt}}
        \toprule
        & {\footnotesize CNP} & {\footnotesize AttCNP} & {\footnotesize TNP} & {\footnotesize TE-TNP} & {\footnotesize ConvCNP} & {\footnotesize SConvCNP} \\
        \midrule
        {\footnotesize Number of parameters (million)}  
            & $1.3$  & $2.1$ & $2.6$ & $3.1$ & $3.8$ & $3.7$ \\
        \bottomrule
    \end{tabular}}
\end{table*}

\paragraph{Forward run time.}
Table~\ref{table: synthetic-regression-forward-run-time} reports the forward-pass runtime (in seconds) for a batch of 16 tasks during both training and validation across all models evaluated in our 1D synthetic regression experiments. As described in Appendix~\ref{appendix: synthetic-data-experiment-setup}, the number of query points is fixed to 256 during validation, whereas during training it is sampled uniformly from $\mathcal{U}[5, 25)$.
\begin{table*}[!ht]\centering
    \caption{Average forward-pass runtime (in seconds) for a batch of 16 tasks during training and validation across all models in the 1D synthetic regression setting.}
    \vspace{2.5pt}
    \renewcommand{\arraystretch}{1.3}
    \label{table: synthetic-regression-forward-run-time}
    \scalebox{1.0}{
    \begin{tabular}{l@{\hskip 15pt} c@{\hskip 10pt} c@{\hskip 10pt} c@{\hskip 10pt} c@{\hskip 10pt} c@{\hskip 10pt} c@{\hskip 1pt}}
        \toprule
        & {\footnotesize CNP} & {\footnotesize AttCNP} & {\footnotesize TNP} & {\footnotesize TE-TNP} & {\footnotesize ConvCNP} & {\footnotesize SConvCNP} \\
        \midrule
        {\footnotesize Train}     & $0.003$ & $0.004$  & $0.007$ & $0.014$ & $0.009$ & $0.009$\\
        {\footnotesize Validation}   & $0.006$ & $0.009$  & $0.013$ & $0.023$ & $0.033$ & $0.030$ \\
        \bottomrule
    \end{tabular}}
\end{table*}

\subsubsection{Data and Experimental Setup}
\label{appendix: synthetic-data-experiment-setup}

We evaluate our methods on four families of synthetic 1D regression tasks, each defined by a distinct stochastic generative process: a Gaussian process (GP) with a Matérn--5/2 kernel, a GP with a periodic kernel, a sawtooth-wave generator, and a square-wave generator. GP tasks are sampled using \texttt{GPyTorch} \citep{gardner2018gpytorch}. All processes include independent Gaussian observation noise.

\paragraph{Generative Processes.} For each task family, process-specific hyperparameters are sampled independently at the task level.

\begin{itemize}
    \item \textbf{GP with Matérn--5/2 kernel.}
    Functions are sampled from $f \sim \mathcal{GP}(0, k_{\text{m5/2}} + \sigma_0^2 I)$, where
    \begin{equation*}
        k_{\text{m5/2}}(x, x') = \frac{2^{-1.5}}{\Gamma(2.5)} \, (\frac{\sqrt{5}}{\lambda}|x - x'|)^{2.5} K_{2.5}(\frac{\sqrt{5}}{\lambda}|x - x'|)
    \end{equation*}
    and $K_{2.5}$ is the modified Bessel function of the second kind. The lengthscale is sampled as $\lambda \sim \mathcal{U}[0.25, 1)$. Observation noise standard deviation is $\sigma_0 = 0.1$.

    \item \textbf{GP with periodic kernel.}
    Functions are sampled from $f \sim \mathcal{GP}(0, k_{\text{p}} + \sigma_0^2 I)$, where
    \begin{equation*}
        k_{\text{p}}(x, x') = \exp\!\left( -\frac{2\sin^2(\pi|x - x'|/\rho)}{\lambda^2} \right)
    \end{equation*}
    with period $\rho \sim \mathcal{U}[0.5,2)$ and lengthscale $\lambda \sim \mathcal{U}[0.25,1)$. Observation noise standard deviation is $\sigma_0 = 0.1$.

    \item \textbf{Sawtooth wave.}
    Functions are sampled from $f \sim \mathcal{GP}(m_{\text{saw}}, \sigma_0^2 I)$, where the mean function is
    \begin{equation*}
        m_{\text{saw}}(x) = 2 \left( \left(\xi \left(ux - c\right)\right) \bmod 1 \right) - 1
    \end{equation*}
    with frequency $\xi \sim \mathcal{U}[0.5,5)$, direction $u \in \{+1,-1\}$ (sampled uniformly), and phase offset $c \sim \mathcal{U}[0,1)$. Observation noise standard deviation is $\sigma_0 = 0.05$.
    
    \item \textbf{Square wave.}
    Functions are sampled from $f \sim \mathcal{GP}(m_{\text{sq}}, \sigma_0^2 I)$, where the mean function is
    \begin{equation*}
        m_{\text{sq}}(x) = 2 \, \mathds{1}_{\{ ((\xi x - c) \bmod 1) < D \}} - 1
    \end{equation*}
    with frequency $\xi \sim \mathcal{U}[0.5,5)$, duty cycle $D \sim \mathcal{U}[0.25,0.75)$, and phase offset $c \sim \mathcal{U}[0,1)$. Observation noise standard deviation is $\sigma_0 = 0.05$.
    
\end{itemize}

Each model is trained for 500 epochs using AdamW~\citep{loshchilov2017decoupled} with a learning rate of $5\times 10^{-4}$. We apply gradient clipping with a maximum norm of $0.5$ \citep{pascanu2013difficulty}. Each epoch consists of 1000 iterations, and every iteration processes a batch of 16 tasks, yielding a total of 8 million on-the-fly sampled training tasks. For each batch, the numbers of context and query points are independently drawn as 
$n_c \sim \mathcal{U}[5, 25)$ and $n_q \sim \mathcal{U}[5, 25)$. These values are shared across all tasks in the batch. Input locations are sampled uniformly and independently from $[-3, 3)$ for each task.

For validation, we use a fixed set of 4{,}096 tasks, organized into 256 batches of 16 tasks. In these tasks, the number of query points is fixed at 256, while the number of context points is sampled using the same procedure as during training. Testing follows the same configuration as validation, except that we evaluate on 16{,}000 test tasks. Unlike the dynamically generated training tasks, the validation and test sets remain fixed across all experiments and runs.

\subsection{Predator-Prey Model}\label{appendix: pred–prey-experiment-details}

\subsubsection{Model Architectures}\label{appendix: pred–prey-model-architectures}
The model architectures and parameter counts closely follow those described in
\ref{appendix: synthetic-regression-model-architectures}. The only modification concerns the discretization of the functional embedding in the ConvCNP and the SConvCNP. Specifically, we first expand the range defined by the minima and maxima of all context and query coordinates by a margin of 0.5 in each dimension. The resulting interval is then uniformly discretized at a resolution of 48 points per unit length.

\subsubsection{Data and Experimental Setup}\label{appendix: pred–prey-data-generation}
We generate data from a stochastic variant of the Lotka--Volterra predator--prey system \citep{Lotka1910contribution, Volterra1926variazioni}, following the formulation of \citet{bruinsma2023autoregressive}. Let $U_t$ and $V_t$ denote the prey and predator populations at time $t$. Their dynamics evolve according to
\begin{align*}
    dU_t &= \alpha U_t\,dt - \beta U_t V_t\,dt 
           + \sigma U^{\nu}_t\, dB^{(1)}_t, \\
    dV_t &= -\gamma U_t\,dt + \delta U_t V_t\,dt 
           + \sigma V^{\nu}_t\, dB^{(2)}_t ,
\end{align*}
where $B^{(1)}_t$ and $B^{(2)}_t$ are independent Brownian motions. 
The deterministic drift recovers classical predator--prey behavior: prey grow exponentially at rate $\alpha$ in the absence of predators, while predators decline at rate $\gamma$ without prey. The interaction terms $\beta U_t V_t$ and $\delta U_t V_t$ model predation and predator reproduction. Stochasticity enters through the multiplicative noise terms $\sigma U^{\nu}_t$ and $\sigma V^{\nu}_t$, where $\sigma$ controls noise intensity and $\nu$ determines how fluctuations scale with population size. Setting $\nu = 1$ yields noise proportional to population level, while $\nu < 1$ and $\nu > 1$ induce sublinear and superlinear scaling, respectively.

\begin{table*}[!ht]\centering
    \caption{Parameter distributions for the stochastic Lotka--Volterra equations.}
    \vspace{2.5pt}
    \renewcommand{\arraystretch}{1.2}
    \label{table: stoch-Lotka-Volterra-parameters}
    \scalebox{1.0}{
    \begin{tabular}{l@{\hskip 15pt}l}
        \toprule
        Parameter & Distribution \\
        \midrule
        Initial condition $U_{-10}$ & $\mathcal{U}([5, 100])$ \\ 
        Initial condition $V_{-10}$ & $\mathcal{U}([5, 100])$ \\
        $\alpha$ & $\mathcal{U}([1.0, 5.0])$ \\
        $\beta$ & $\mathcal{U}([0.04, 0.08])$ \\
        $\gamma$ & $\mathcal{U}([1.0, 2.0])$ \\
        $\delta$ & $\mathcal{U}([0.04, 0.08])$ \\
        $\sigma$ & $\mathcal{U}([0.5, 10])$ \\
        $\eta$ & $\mathcal{U}([1, 5])$ \\
        $\nu$ & Fixed at $1/6$ \\
        \bottomrule
    \end{tabular}}
\end{table*}

We simulate trajectories using the parameter distributions in Table~\ref{table: stoch-Lotka-Volterra-parameters} over a total of 110 years, discarding the first 10 years as burn-in to reduce sensitivity to initial conditions. Integration is performed using the Euler--Maruyama method with time step $\Delta t = 0.022$, producing 5000 solver steps over the interval $t \in [-10, 100)$. Although the solver runs at this finer resolution, we record states only on a uniform grid with spacing $0.05$, yielding approximately 2200 selected time points spanning $[-10, 100)$.

To construct a task from a simulated trajectory, input locations $t$ are sampled uniformly and independently from the recorded time points in $[0,100)$ and paired with the corresponding values $(U_t, V_t)$. We rescale time by a factor of~0.1, mapping the 100-year interval to $[0,10)$, and rescale population sizes by multiplying the values by~0.01.

Models are trained for 500 epochs using AdamW with a learning rate of $10^{-4}$. We apply gradient clipping with a maximum norm of $0.5$. Each epoch consists of 1{,}000 iterations, each processing a batch of 16 tasks, yielding a total of 8 million on-the-fly sampled training tasks. For every batch, the numbers of context and query points are drawn independently as 
$n_c \sim \mathcal{U}[5, 25)$ and $n_q \sim \mathcal{U}[5, 25)$; the sampled values are shared across all tasks within the batch.

For validation, we use a fixed set of 4{,}096 tasks, arranged into 256 batches of 16 tasks. In these tasks, the number of query points is fixed at 256, while the number of context points is sampled using the same procedure as in training. Testing follows the same protocol as validation, except that we evaluate on 16{,}000 test tasks. Unlike the dynamically generated training tasks, the randomization used to construct the validation and test tasks is fixed across all experiments and runs.

\subsection{Traffic Flow}\label{appendix: traffic-flow-experiment-details}
\subsubsection{Model Architectures}\label{appendix: traffic-flow-model-architectures}
The model architectures follow those described in
Section~\ref{appendix: synthetic-regression-model-architectures}.
For the discretization of the functional embedding in both the ConvCNP and the SConvCNP, we use an expanded margin of~0.5 and a resolution of
48~grid points per unit length.

\subsubsection{Data and Experimental Setup}\label{appendix: traffic-flow-data-generation}
We use the California traffic-flow dataset from \textsc{LargeST}~\citep{liu2023largest}, a large-scale benchmark for traffic forecasting, available at \url{https://www.kaggle.com/datasets/liuxu77/largest}. The dataset contains traffic-flow measurements from 8{,}600 loop-detector sensors deployed across California’s highway network, collected at 5-minute intervals between 2017 and 2021. In our experiments, we restrict attention to data from the year~2020.

As a preprocessing step, we discard sensors with more than 50\% missing values. For the remaining sensors, missing entries are filled via linear interpolation between observed measurements, with leading and trailing gaps completed using forward and backward propagation, respectively. Sensors are then randomly partitioned into training, validation, and test sets using a 6{:}1{:}3 split.

Each sensor’s time series is segmented into non-overlapping, continuous 14-day windows (4{,}032 time steps), yielding 26 windows per sensor. Within each window, timestamps are reset to start at~0 and increase in 5-minute increments. We subsequently downsample each window by a factor of~6, resulting in a temporal resolution of 30~minutes. Time indices are rescaled to units of days by dividing by $(60 \times 24)$. Finally, traffic-flow values are normalized to the $[0,1]$ range using min--max statistics computed from the training set.

The processed dataset contains:
\begin{itemize}
    \item \textbf{Training:} 5{,}119 sensors $\rightarrow$ 133{,}094 windows
    \item \textbf{Validation:} 853 sensors $\rightarrow$ 22{,}178 windows
    \item \textbf{Test:} 2{,}561 sensors $\rightarrow$ 66{,}586 windows
\end{itemize}

Models are trained for 100 epochs using AdamW with a learning rate of $10^{-4}$. We apply gradient clipping with a maximum norm of $0.5$. Each epoch consists of approximately 4{,}160 iterations, each processing a batch of 32 tasks. For every batch, the numbers of context and query points are drawn independently as 
$n_c \sim \mathcal{U}[5, 25)$ and $n_q \sim \mathcal{U}[5, 25)$; the sampled values are shared across all tasks within the batch. A task is formed by independently and uniformly sampling $n_c + n_q$ time steps from a window without replacement. Training tasks are sampled on-the-fly from training windows. For validation and testing, we pre-generate fixed sets of tasks:
\begin{itemize}
    \item \textbf{Validation:} 22{,}178 tasks (batched in 32), with $n_q = 256$ and $n_c$ sampled as in training.
    \item \textbf{Test:} 66{,}586 tasks (batched in 32), with $n_q = 50$ and $n_c$ sampled as in training.
\end{itemize}

\subsection{Image Completion}\label{appendix: image-completion-experiment-details}
\subsubsection{Model Architectures}\label{appendix: image-completion-model-architectures}

All models output Gaussian predictive distributions parameterized by a mean and a pre-softplus standard deviation. Because pixel values are normalized to $[0, 1]$, the predicted mean is passed through a sigmoid. The predicted standard deviation is obtained by applying a softplus to the network output, scaling the result by $0.99$, and adding a minimum noise of $0.01$.

\paragraph{Conditional Neural Process (CNP).}
The CNP architecture follows the design used in previous experiments (see Section~\ref{appendix: synthetic-regression-model-architectures}), with the only modification being an increase in the MLP hidden size from $256$ to $512$.

\paragraph{Attentive Conditional Neural Process (AttCNP).}
The AttCNP architecture is likewise based on the design used in earlier experiments (see Section~\ref{appendix: synthetic-regression-model-architectures}). The MLP hidden size is increased from $256$ to $512$. For the self-attention applied to the set of context embeddings, we increase the per-head dimension from $32$ to $64$, set the input/output dimension of each attention layer to $512$, and increase the feedforward subnetwork width from $256$ to $512$. The same modifications apply to the multi-head cross-attention used to update query embeddings (eight heads of dimension~$64$ and feedforward width~$512$).

\paragraph{Transformer Neural Process (TNP).}
As with the other models, all MLP hidden sizes are increased from $256$ to $512$. We adopt the Efficient Query TNP design described in Section~\ref{appendix: synthetic-regression-model-architectures}, which employs a two-branch structure in which context embeddings are processed twice and query embeddings once. In contrast to previous sections, we allocate \emph{separate} parameters for each branch: at each layer, the multi-head self-attention over contexts and the multi-head cross-attention between queries and contexts no longer share parameters. For each multi-head attention module, the feedforward width is increased from $256$ to $512$, and all input/output embedding dimensions are set to $512$. All other configurations follow Section~\ref{appendix: synthetic-regression-model-architectures}.

\paragraph{Convolutional Conditional Neural Process (ConvCNP).}
Because the data lie on a regular grid, we use the on-the-grid implementation of the ConvCNP \citep{gordon2019convolutional}, eliminating the discretization step and improving computational efficiency. Let $\mathrm{I}$ denote an incomplete image, with unobserved pixels filled with dummy values, and let $\mathrm{M}_{c}$ be the binary mask indicating observed (context) pixels. For multi-channel images, the mask is broadcast along the channel dimension.

As in Section~\ref{appendix: synthetic-regression-model-architectures}, the output of the convolutional deep set module has two components: (i) a density channel capturing the spatial distribution of context pixels, and (ii) a kernel-smoothed representation. The kernel is implemented via a 2D convolutional layer (kernel size~11, $d_y$ input channels, 128 output channels, no bias) with a nonnegativity constraint enforced by taking absolute values of the learned weights during the forward pass \citep{gordon2019convolutional}.

The density channel is obtained by convolving this modified filter with the mask $\mathrm{M}_{c}$. The kernel-smoothed component is computed by first multiplying $\mathrm{I}$ elementwise with $\mathrm{M}_{c}$ (setting non-context pixels to zero) and then applying the same modified convolution. Following \citet{gordon2019convolutional}, we omit normalization by the density channel, as it did not provide empirical benefits and occasionally introduced instability; however, we retain the positivity constraint on the kernel.

The resulting representation is processed pointwise by an MLP with two hidden layers of width~128. These features are then fed into a ResNet-style CNN \citep{he2016deep} consisting of six residual convolutional blocks (kernel size~11, 128 channels), following the implementation of \citet{bruinsma2023autoregressive}. The embeddings corresponding to query pixels are finally passed to a decoder MLP with two hidden layers of width~128.

\paragraph{Spectral Convolutional Conditional Neural Process (SConvCNP).}
The SConvCNP mirrors the on-the-grid ConvCNP architecture described above, with the distinction that the ResNet backbone is replaced by a U-shaped FNO. To construct positional information, we uniformly discretize the interval $[-1, 1]$ along each spatial axis into grids whose sizes match the corresponding image dimensions. Their Cartesian product yields the 2D positional encoding, which is concatenated with the output of the convolutional deep set. This concatenated tensor is then passed pointwise through an MLP with one hidden layer of width~128 and GELU activations.

Let $s_h$ and $s_w$ denote the image height and width, respectively. Using the notation introduced in Section~\ref{appendix: synthetic-regression-model-architectures}, the operator consists of the following sequence of residual Fourier blocks:
\begin{itemize}
    \item $\mathrm{L_1} = \mathrm{F}(512,\,512,\,(s_h, s_w),\,(s_h, s_w),\,32)$
    \item $\mathrm{L_2} = \mathrm{F}(512,\,512,\,(s_h, s_w),\,(\lfloor s_h/4 \rfloor,\, \lfloor s_w/4 \rfloor),\,8)$
    \item $\mathrm{L_3} = \mathrm{F}(512,\,512,\,(\lfloor s_h/4 \rfloor,\, \lfloor s_w/4 \rfloor),\,(\lfloor s_h/16 \rfloor,\, \lfloor s_w/16 \rfloor),\,2)$
    \item $\mathrm{L_4} = \mathrm{F}(512,\,512,\,(\lfloor s_h/16 \rfloor,\, \lfloor s_w/16 \rfloor),\,(\lfloor s_h/16 \rfloor,\, \lfloor s_w/16 \rfloor),\,2)$
    \item $\mathrm{L_5} = \mathrm{F}(512,\,512,\,(\lfloor s_h/16 \rfloor,\, \lfloor s_w/16 \rfloor),\,(\lfloor s_h/4 \rfloor,\, \lfloor s_w/4 \rfloor),\,8)$
    \item $\mathrm{L_6} = \mathrm{F}(512,\,512,\,(\lfloor s_h/4 \rfloor,\, \lfloor s_w/4 \rfloor),\,(s_h, s_w),\,16)$
    \item $\mathrm{L_7} = \mathrm{F}(512,\,512,\,(s_h, s_w),\,(s_h, s_w),\,32)$
\end{itemize}

The forward pass follows a standard U-shaped pattern. Layers $\mathrm{L_1} \rightarrow \mathrm{L_2} \rightarrow \mathrm{L_3} \rightarrow \mathrm{L_4}$ are applied sequentially. On the upward path, $\mathrm{L_5}$ receives the channel-wise concatenation of the outputs of $\mathrm{L_4}$ and $\mathrm{L_3}$. Similarly, $\mathrm{L_6}$ receives the concatenation of the outputs of $\mathrm{L_5}$ and $\mathrm{L_2}$, and $\mathrm{L_7}$ receives the concatenation of the outputs of $\mathrm{L_6}$ and $\mathrm{L_1}$. The output of $\mathrm{L_7}$ is then concatenated with the features from the initial MLP and fed into a final MLP (one hidden layer, width~128, GELU) to yield the SConvCNP representation consumed by the decoder.

\paragraph{Parameter Count.}
Table \ref{table: image-completion-parameter-count} summarizes the number of learnable parameters for all models evaluated in our image completion experiments. 

\begin{table*}[!ht]\centering
    \caption{Learnable parameter counts for all models evaluated in the image completion experiments.}
    \vspace{2.5pt}
    \renewcommand{\arraystretch}{1.3}
    \label{table: image-completion-parameter-count}
    \scalebox{1.0}{
    \begin{tabular}{l@{\hskip 15pt} c@{\hskip 10pt} c@{\hskip 10pt} c@{\hskip 10pt} c@{\hskip 10pt} c@{\hskip 1pt}}
        \toprule
        & {\footnotesize CNP} & {\footnotesize AttCNP} & {\footnotesize TNP} & {\footnotesize ConvCNP} & {\footnotesize SConvCNP} \\
        \midrule
        {\footnotesize Number of parameters (million)}  
            & $5$  & $8.4$ & $13.7$  & $12.2$ & $14.2$ \\
        \bottomrule
    \end{tabular}}
\end{table*}

\subsubsection{Data and Experimental Setup}\label{appendix:image-completion-data}
For our image-completion experiments, we use the Describable Textures Dataset (DTD; \citet{cimpoi2014describing}). We adopt the standard DTD split, which contains 1{,}880 images each for training, validation, and testing. All images are RGB, with heights ranging from 231 to 778 pixels and widths from 271 to 900 pixels. During training, validation, and testing, we sample from each image a random $192 \times 192$ crop, which we then downsample to $64 \times 64$. Pixel coordinates of the resulting $64 \times 64$ grid along each axis are linearly mapped to $[-1, 1]$, and pixel intensities are normalized to $[0, 1]$ independently across channels.

Models are trained for 500 epochs using AdamW with a learning rate of $10^{-4}$. We apply gradient clipping with a maximum norm of $0.5$. At each epoch, the model processes batches of 16 tasks. For each batch, we draw the number of context pixels as $n_c \sim \mathcal{U}[5, 1024)$, and treat all remaining pixels as queries, i.e. $n_q = 64 \times 64 - n_c$. A single sampled value of $n_c$ is shared across all tasks within the batch.

\subsection{Ablation Studies}\label{appendix: ablation}
We conduct a series of ablation studies to assess the contribution of three core design choices in the SConvCNP:
\begin{enumerate}
    \item the number of retained Fourier modes,
    \item the discretization resolution of the functional embedding, and
    \item the use of positional encodings.
\end{enumerate}
Experiments are conducted on two families of one-dimensional functions: samples drawn from a GP with a Matérn--$5/2$ kernel, and a deterministic sawtooth waveform. Training follows the protocol described in Section~\ref{appendix: synthetic-regression-experiment-details}, with the only difference being the use of a smaller SConvCNP model (3.1M parameters, 32 Fourier modes). Results are reported over 8{,}096 test tasks, and the main findings are summarized below.

\paragraph{Number of Fourier modes.}
\begin{table*}[!h]\centering
    \caption{Predictive log-likelihood of SConvCNP for different numbers of Fourier modes $m$.}
    \vspace{2.5pt}
    \renewcommand{\arraystretch}{1.2}
    \label{table: 1d-synthetic-ablation-results-number-of-modes}
    \scalebox{1.0}{
    \begin{tabular}{l@{\hskip 15pt} c@{\hskip 10pt} c@{\hskip 10pt} c@{}}
        \toprule
            & {\footnotesize $m = 8$} & {\footnotesize $m = 16$} & {\footnotesize $m = 32$} \\
        \midrule
            {\footnotesize Mat\'ern 5/2}     & ${-0.29_{\pm0.00}}$ & ${-0.29_{\pm0.00}}$  & ${-0.29_{\pm0.00}}$ \\
            {\footnotesize Sawtooth wave}   & ${-0.14_{\pm0.03}}$ & ${0.20_{\pm0.02}}$  & ${0.80_{\pm0.03}}$ \\
        \bottomrule
    \end{tabular}}
\end{table*}
Increasing the number of retained Fourier modes yields substantial improvements for the sawtooth signal, while having a negligible effect on Matérn--$5/2$ functions (Table~\ref{table: 1d-synthetic-ablation-results-number-of-modes}). This behavior aligns with the spectral properties of the underlying functions. The sawtooth wave exhibits Fourier amplitudes that decay as $1/\xi$, leaving significant energy at high frequencies and necessitating a large number of modes for accurate reconstruction.
In contrast, Matérn--$5/2$ samples have power spectra that decay as $1/\xi^{6}$, resulting in extremely weak high-frequency content; consequently, a small set of low-frequency modes suffices to capture nearly all signal energy.

\paragraph{Discretization resolution.}
\begin{table*}[!h]\centering
    \caption{Predictive log-likelihood of SConvCNP for different discretization resolutions (number of points per unit).}
    \vspace{2.5pt}
\renewcommand{\arraystretch}{1.2}
    \label{table: 1d-synthetic-ablation-results-discretization-resolution}
    \scalebox{1.0}{
    \begin{tabular}{l@{\hskip 15pt} c@{\hskip 10pt} c@{\hskip 10pt} c@{}}
    \toprule
        & {\footnotesize 16} & {\footnotesize 32} & {\footnotesize 64} \\
        \midrule
        {\footnotesize Mat\'ern 5/2}     & ${-0.30_{\pm0.00}}$ & ${-0.29_{\pm0.00}}$  & ${-0.29_{\pm0.00}}$ \\
        {\footnotesize Sawtooth wave}   & ${0.03_{\pm0.02}}$ & ${0.26_{\pm0.08}}$  & ${0.20_{\pm0.02}}$ \\
    \bottomrule
    \end{tabular}}
\end{table*}
A finer discretization improves performance for the sawtooth signal but has little impact on Matérn--$5/2$ samples (Table~\ref{table: 1d-synthetic-ablation-results-discretization-resolution}). The slowly decaying spectrum of the sawtooth implies a high effective Nyquist frequency, requiring dense sampling to resolve sharp transitions. By contrast, the spectral mass of Matérn--$5/2$ functions is concentrated at lower frequencies, so coarser discretizations can adequately capture the relevant structure.

\paragraph{Positional encoding.}
\begin{table*}[!h]\centering
    \caption{Predictive log-likelihood of SConvCNP with and without positional encoding.}
    \vspace{2.5pt}
    \renewcommand{\arraystretch}{1.2}
    \label{table: 1d-synthetic-ablation-results-positional-encoding}
    \scalebox{1.0}{
    \begin{tabular}{l@{\hskip 15pt} c@{\hskip 10pt} c@{}}
        \toprule
        & {\footnotesize with positional encoding} & {\footnotesize without positional encoding} \\
        \midrule
        {\footnotesize Mat\'ern 5/2}     & ${-0.29_{\pm0.00}}$ & ${-0.31_{\pm0.00}}$ \\
        {\footnotesize Sawtooth wave}   & ${0.80_{\pm0.03}}$ & ${0.67_{\pm0.02}}$ \\
    \bottomrule
    \end{tabular}}
\end{table*}
Incorporating positional encodings consistently improves predictive performance, with particularly pronounced gains on the sawtooth tasks (Table~\ref{table: 1d-synthetic-ablation-results-positional-encoding}). This suggests that explicit location information helps disambiguate high-frequency and non-smooth patterns that are not fully captured by fully translation equivariant features alone in the SConvCNP.

\newpage
\section*{NeurIPS Paper Checklist}

\begin{enumerate}
\item {\bf Claims}
    \item[] Question: Do the main claims made in the abstract and introduction accurately reflect the paper's contributions and scope?
    \item[] Answer: \answerYes{}
    \item[] Justification: The abstract and introduction clearly state that we introduce Spectral Convolutional Conditional Neural Processes (SConvCNPs), explain the motivation for using spectral methods to address limitations in ConvCNPs, and accurately describe the scope of our experimental validation. The claims match the actual content and results presented in the paper.

\item {\bf Limitations}
    \item[] Question: Does the paper discuss the limitations of the work performed by the authors?
    \item[] Answer: \answerYes{}
    \item[] Justification: Section 3.1 discusses the limitations of our approach, including the trade-off between positional encodings and translation equivariance, and the discretization sensitivity of the FFT.

\item {\bf Theory assumptions and proofs}
    \item[] Question: For each theoretical result, does the paper provide the full set of assumptions and a complete (and correct) proof?
    \item[] Answer: \answerNA{}
    \item[] Justification: This paper primarily presents a methodological contribution with empirical validation rather than theoretical results requiring formal proofs.

\item {\bf Experimental result reproducibility}
    \item[] Question: Does the paper fully disclose all the information needed to reproduce the main experimental results of the paper to the extent that it affects the main claims and/or conclusions of the paper (regardless of whether the code and data are provided or not)?
    \item[] Answer: \answerYes{}
    \item[] Justification: The paper provides detailed descriptions of the model architecture, data generation process, and experimental setup, ensuring that the main results can be reproduced. In addition, all implementations are publicly available in our code repository.

\item {\bf Open access to data and code}
    \item[] Question: Does the paper provide open access to the data and code, with sufficient instructions to faithfully reproduce the main experimental results, as described in supplemental material?
    \item[] Answer: \answerYes{}
    \item[] Justification: All implementations are publicly available in our code repository.

\item {\bf Experimental setting/details}
    \item[] Question: Does the paper specify all the training and test details (e.g., data splits, hyperparameters, how they were chosen, type of optimizer, etc.) necessary to understand the results?
    \item[] Answer: \answerYes{}
    \item[] Justification: The paper provides detailed descriptions of the model architecture, data generation process, and experimental setup

\item {\bf Experiment statistical significance}
    \item[] Question: Does the paper report error bars suitably and correctly defined or other appropriate information about the statistical significance of the experiments?
    \item[] Answer: \answerYes{}
    \item[] Justification: All experimental results in Tables 1-3 report standard deviations across 4 seeds, providing a clear measure of the statistical variability in our results.

\item {\bf Experiments compute resources}
    \item[] Question: For each experiment, does the paper provide sufficient information on the computer resources (type of compute workers, memory, time of execution) needed to reproduce the experiments?
    \item[] Answer: \answerYes{}
    \item[] Justification:The paper explicitly states that all computations were performed on a single NVIDIA A100 GPU with 40 GB of memory. Additionally, forward-pass runtimes for all models on one of the benchmarks are reported, providing further clarity on the execution time requirements.

\item {\bf Code of ethics}
    \item[] Question: Does the research conducted in the paper conform, in every respect, with the NeurIPS Code of Ethics \url{https://neurips.cc/public/EthicsGuidelines}?
    \item[] Answer: \answerYes{}
    \item[] Justification: The research centers on methodological advancements in neural network architectures for probabilistic meta-learning, relying solely on synthetic data and publicly available datasets. The work does not involve human subjects, sensitive data, or applications with foreseeable harmful impact, and it faithfully represents all results. Accordingly, it aligns with the NeurIPS Code of Ethics.

\item {\bf Broader impacts}
    \item[] Question: Does the paper discuss both potential positive societal impacts and negative societal impacts of the work performed?
    \item[] Answer: \answerNo{}
    \item[] Justification: The paper does not currently include a dedicated discussion of the broader societal impacts of the work, and therefore does not address potential positive or negative implications.

\item {\bf Safeguards}
    \item[] Question: Does the paper describe safeguards that have been put in place for responsible release of data or models that have a high risk for misuse (e.g., pretrained language models, image generators, or scraped datasets)?
    \item[] Answer: \answerNA{}
    \item[] Justification: Our work focuses on meta-learning for regression tasks. The models and data used do not pose significant risks for misuse or dual-use concerns.

\item {\bf Licenses for existing assets}
    \item[] Question: Are the creators or original owners of assets (e.g., code, data, models), used in the paper, properly credited and are the license and terms of use explicitly mentioned and properly respected?
    \item[] Answer: \answerYes{}
    \item[] Justification: We have been meticulous in citing all prior work and assets used in our experiments. For each external codebase, dataset, or model, we provide proper attribution and ensure that the associated licenses and terms of use are respected.

\item {\bf New assets}
    \item[] Question: Are new assets introduced in the paper well documented and is the documentation provided alongside the assets?
    \item[] Answer: \answerYes{}
    \item[] Justification: The paper provides a detailed description of the experiments, and the complete code implementation is publicly released with accompanying documentation, ensuring that all newly introduced assets are well described and accessible.

\item {\bf Crowdsourcing and research with human subjects}
    \item[] Question: For crowdsourcing experiments and research with human subjects, does the paper include the full text of instructions given to participants and screenshots, if applicable, as well as details about compensation (if any)? 
    \item[] Answer: \answerNA{}
    \item[] Justification: This research does not involve crowdsourcing or human subjects.

\item {\bf Institutional review board (IRB) approvals or equivalent for research with human subjects}
    \item[] Question: Does the paper describe potential risks incurred by study participants, whether such risks were disclosed to the subjects, and whether Institutional Review Board (IRB) approvals (or an equivalent approval/review based on the requirements of your country or institution) were obtained?
    \item[] Answer: \answerNA{}
    \item[] Justification: This research does not involve human subjects, so IRB approval was not required.

\item {\bf Declaration of LLM usage}
    \item[] Question: Does the paper describe the usage of LLMs if it is an important, original, or non-standard component of the core methods in this research? Note that if the LLM is used only for writing, editing, or formatting purposes and does not impact the core methodology, scientific rigorousness, or originality of the research, declaration is not required.
    \item[] Answer: \answerNA{}
    \item[] Justification: The research methodology does not involve the use of large language models as components of the proposed method. Any LLM assistance used was solely for writing and editing purposes and does not impact the scientific rigor or originality of the research.

\end{enumerate}

\end{document}